\documentclass[runningheads]{llncs}
\usepackage{graphicx}

\usepackage{tikz}
\usepackage{comment}
\usepackage{amsmath,amssymb} 
 \usepackage{color}
\usepackage{epsfig}
\usepackage{hyperref}
\usepackage{tabulary} 
\usepackage{tabularx} 
\usepackage{multirow}
\usepackage{hhline} 
\usepackage{array} 
\usepackage{subfiles} 
\usepackage{adjustbox}
\usepackage[tight,normalsize,sf,SF]{subfigure}

\hypersetup{
     colorlinks = true,
     linkcolor = red,
     anchorcolor = red,
     citecolor = green,
     filecolor = blue,
     urlcolor = magenta
     }

\begin{document}
\pagestyle{headings}
\mainmatter

\title{Mitigating Dataset Imbalance via Joint Generation and Classification}

\title{Mitigating Dataset Imbalance via Joint Generation and Classification}
\titlerunning{Mitigating Dataset Imbalance via Joint Generation and Classification}

\author{\hspace{-6mm}Aadarsh Sahoo\inst{1}\thanks{The first two authors contributed equally. The work was done when Ankit Singh was at IIT Kharagpur.} \and
Ankit Singh\inst{2}\textsuperscript{*} \and
Rameswar Panda\inst{3} \and Rogerio Feris\inst{3} \and Abir Das\inst{1}}
\authorrunning{A. Sahoo, A. Singh, R. Panda, R. Feris and A. Das}

\institute{IIT Kharagpur, West Bengal, India \and IIT Madras, Tamil Nadu, India \and MIT-IBM Watson AI Lab, Massachusetts, USA}

\maketitle

\begin{abstract}
Supervised deep learning methods are enjoying enormous success in many practical applications of computer vision and have the potential to revolutionize robotics.
However, the marked performance degradation to biases and imbalanced data questions the reliability of these methods.
In this work we address these questions from the perspective of dataset imbalance resulting out of severe under-representation of annotated training data for certain classes and its effect on both deep classification and generation methods.
We introduce a joint dataset repairment strategy by combining a neural network classifier with Generative Adversarial Networks (GAN) that makes up for the deficit of training examples from the under-representated class by producing additional training examples. We show that the combined training helps to improve the robustness of both the classifier and the GAN against severe class imbalance. We show the effectiveness of our proposed approach on three very different datasets with different degrees of imbalance in them. The code is available at \color{magenta}{\url{https://github.com/AadSah/ImbalanceCycleGAN}}.
\end{abstract}

\section{Introduction}
\label{sec:intro}
Deep neural networks (DNN) and large-scale datasets have been instrumental behind the remarkable progress in computer vision research.
The data-driven feature and classifier learning enabled Convolutional Neural Networks (CNNs) to achieve superior performance than traditional machine learning methods.
However, the over-reliance on data has brought with it new problems.
One of them is the problem of becoming too adapted to the dataset by essentially memorizing all its idiosyncrasies~\cite{Ponce2006Dataset,Torralba2011Unbiased}.
Due to the presence of massive number of learnable parameters, most DNNs require huge amount of annotated examples for each class.
This is resource consuming, expensive and often impractical too in cases where recognition involves rare classes.
Such a situation is not rare where large number of annotated examples for one class and only a few for the other are available resulting in an imbalanced dataset.

\begin{figure}[t]
	\begin{center}
	\includegraphics[width=0.99\linewidth]{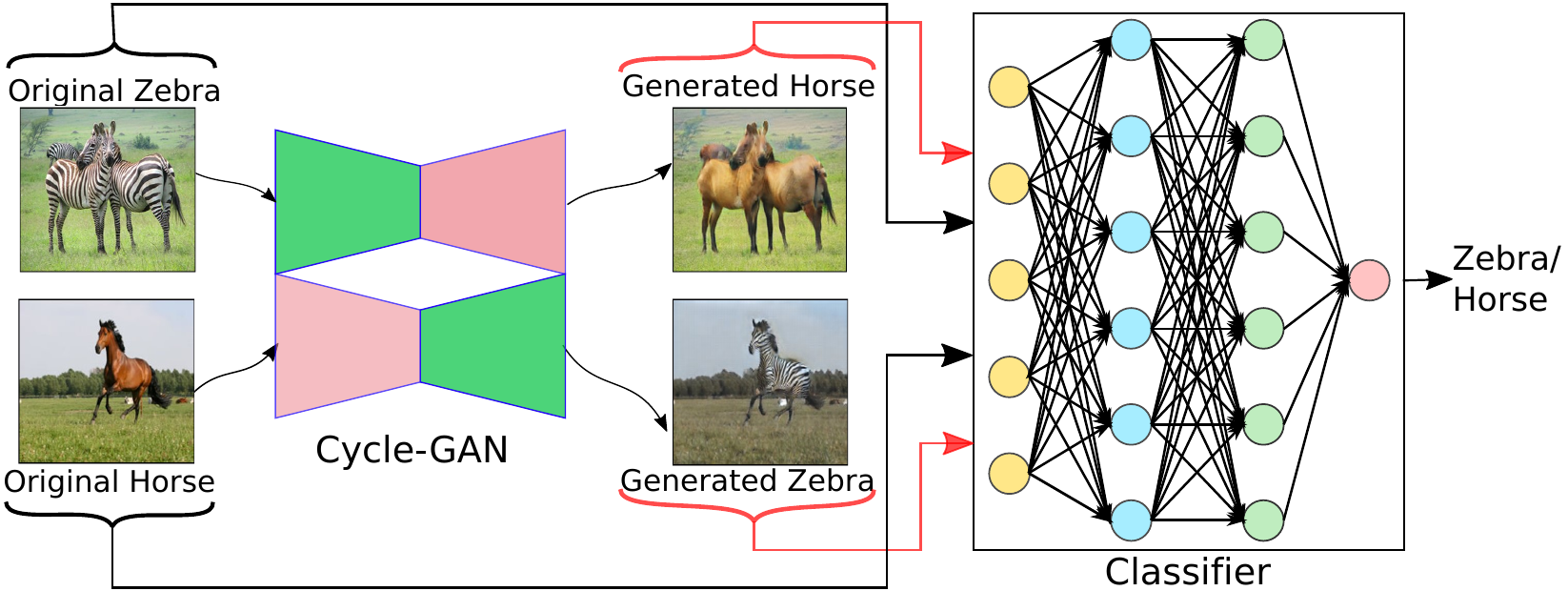}
	\end{center}
	\vskip -0.1in
	\caption{We propose a joint generation and classification network to tackle severe dataset imbalance in classification scenario. The framework employs a Cycle-consistent GAN to generate new training examples from existing ones and feeds both to the classifier. The GAN and the classifier are trained jointly in an alternating fashion. Our model improves the classification performance especially when the dataset is highly imbalanced and the performance of the GAN is either at par with or experiences minor deterioration compared to the case when the classifier is not trained jointly.}
	\label{fig:schematic}
	\vskip -0.2in
\end{figure}

Class imbalance is a classic problem in machine learning and is known to have detrimental effect~\cite{Buda2018Systematic,Haixiang2017Learning,Japkowicz2002Class}.
Classifiers trained traditionally minimizes the average error across all training examples.
Such a strategy often results in fitting to the majority class (\textit{i.e.}, the class with large number of annotated training examples) only.
Simply by virtue of their numbers, fitting to the majority class reduces the overall training error than fitting to the minority class.

Traditionally, dataset imbalance mitigation methods have tried to put all classes in the level playing field by focusing on to take away the number advantage enjoyed by the majority class.
This includes oversampling minority population~\cite{Chawla2002Smote} or undersampling the majority population~\cite{Drummond2003Class,Haixiang2017Learning}, providing more importance to errors from the minority class while training~\cite{Zhou2006Training} or adjusting the predicted class probabilities during inference according to the prior class probabilities~\cite{Richard1991Neural} \textit{etc}.
However, these approaches are limited to the data at hand in the sense that these can not generate additional unseen data that can help increase the much needed diversity for unbiased training.
In this paper, we advocate for a generative approach to address dataset imbalance.
A Generative Adversarial Network (GAN) is used to transform majority class examples to minority class and vice-versa.
Starting from an imbalanced dataset, this generates training examples that are not only much broader and diverse in nature but also balanced in terms of the number of examples from the two different classes.

Our proposed approach explores a Cycle-consistent GAN or simply cycle-GAN~\cite{Zhu2017Unpaired} that is recently proposed to translate image from one domain (source) to another domain (target) (see Fig.~\ref{fig:schematic}).
In addition to traditional generator and discriminator losses during training, a cycle-GAN employs a \emph{cycle-consistency loss} that encourages the learned translations to be ``cycle consistent'', \textit{i.e.}, if an image is translated from a source to a target domain and then it is translated back to the target domain again, original image and the doubly translated image should be same.
We learn a joint architecture by extending cycle-GAN that feeds a classifier with images translated from a source class to a target class where the target class is also the minority class in the imbalanced dataset.
Instead of using the cycle-GAN as only a component of the classification pipeline, we employ a joint training strategy that first trains the classifier using images generated by the cycle-GAN and then trains the cycle-GAN by back-propagating the classifier loss through it.
The loss from the classifier acts as a multi-task supervision
for the cycle-GAN in addition to the cycle-consistency and the traditional GAN loss.
After few such iterations of alternate classifier and GAN training, the GAN learns to generate appropriate images for training a balanced classifier.
Our key insight is that one can use a GAN to alleviate dataset imbalance effect on classifier and at the same time the gradients from the classifier loss can be propagated back to the GAN, to mitigate the same for the GAN.

We evaluate our proposed approach on three different datasets.
The first one is the CelebA dataset~\cite{Liu2015Deep} which is a largescale face dataset created from face images of celebrities, the second one is the Horse2Zebra dataset~\cite{Zhu2017Unpaired}
which is a collection of natural images of horses and zebras downloaded from the internet and the third one is CUB-200-2011 dataset~\cite{WahCUB2011Caltech} consisting of fine-grained images of 200 bird species.
Specifically, we demonstrate that in highly imbalanced scenarios, both classifier and the GAN achieves superior performances than traditional imbalance mitigation approaches.


\section{Related Works}
\label{sec:rel_works}

Class imbalance in machine learning has been studied long~\cite{Japkowicz2002Class}.
However, in the midst of present data revolution, the problem has become ever so important.
While images corresponding to some concepts or classes are available in plenty and easy to annotate, examples corresponding to many concepts are either rare or might require expert annotators.
As a result, many real world datasets are highly imbalanced.
MS-COCO~\cite{Lin2014Microsoft}, SUN database~\cite{Xiao2010Sun}, DeepFashion~\cite{Liu2016Deepfashion}, Places~\cite{Zhou2014Learning}, iNaturalist~\cite{Horn2018The} \text{etc}. are all examples of datasets where the number of images in the most common class and the number of images in the least common class are hugely different.
The community has commonly adopted two different kinds of approaches to address class imbalance: i) dataset level methods and ii) classifier or algorithmic level methods.
A multidimensional taxonomy and categorization of the class imbalance problems can be obtained in the review papers~\cite{Buda2018Systematic,He2008Learning}.

The dataset level approaches consist of modification of imbalanced data in order to provide a balanced distribution with the aim that no change at the algorithm level is necessary and standard training algorithms would work.
These are based on sampling either the majority or the minority classes differently.
Basic random oversampling~\cite{Chawla2002Smote,Estabrooks2004Multiple,Weiss2001Effect} involves replicating randomly selected training examples from the minority class such that the total number of minority and majority class examples that goes to train the classifier is same.
However the same mechanism can be used to achieve varying degrees of class distribution balance by varying the amount of replication.
Oversampling methods are simple and effective, however there are evidences that these can lead to overfitting~\cite{Chawla2002Smote,Wang2014Hybrid}.
To this end, various adaptive and informative oversampling methods have been proposed.
SMOTE~\cite{Chawla2002Smote} creates synthetic examples of the minority class as a convex combination of existing examples and their nearest neighbors.
Borderline-SMOTE~\cite{Han2005Borderline} extends SMOTE by choosing more training examples near the class boundaries.
Authors in~\cite{Batista2004astudy} used a modified condensed nearest neighbor rule~\cite{Tomek1976Two} to sample examples such that all minimally distanced nearest neighbor pairs are of same class and learned a classifier on these samples only.
In CNNs, Shen~\textit{et al.} ~\cite{Shen2016Relay} used a class-aware sampling strategy where the authors first sample a class and then sample an image from that class to form minibatches that are uniform with respect to the classes.

Another variant of sampling is undersampling where training examples are removed randomly from the majority classes until the desired balance in the curated dataset is achieved.
This approach, though counter-intuitive works better than oversampling in some situations~\cite{Drummond2003Class}.
One obvious disadvantage of undersampling is that discarding examples may cause the classifier to miss out on important variabilities in the dataset.
EasyEnsemble
and BalanceCascade~\cite{Liu2008Exploratory} are two variants of informed undersampling that addresses this issue.
EasyEnsemble creates an ensemble of classifiers by independently sampling several subsets from the minority class while BalanceCascade develops the ensemble classifier by systematically selecting the majority class samples.
The one-sided selection~\cite{Kubat1997Addressing} approach discards redundant examples especially close to the boundary to get a more informed reduced sample set than random undersampling of majority class.
Though sampling based methods change the data distribution to be more balanced, they hardly change the data itself.
Our proposed method on the other hand is not only able to change the data distribution to make it more balanced, but also adds variability as additional data is generated in the process.

Among the approaches that modify the learning algorithm without changing the dataset distribution, Zhou \textit{et al.}~\cite{Zhou2006Training} weigh misclassification of the minority class examples more than the majority class examples during training.
A naive approach~\cite{Lawrence1998Neural} is to weigh the predicted probabilities of different classes during inference according to the prior probability of occurrences of these classes in the training set.
Recently Cui~\textit{et al.},~\cite{Cui2019Class} proposed a class balanced loss depending on the ``effective'' number of training examples per class based on the fact that the additional benefit diminishes with increasing number of examples.
This generalizes the class specific loss balancing based on frequency of the samples~\cite{Zhou2006Training} to their effective frequencies.

GAN~\cite{Goodfellow2014_Generative} and its variants use a minimax game to model high dimensional distributions of visual data enabling them to be used for data augmentation~\cite{Ratner2017Learning} as well as for generating new images~\cite{Antoniou2017Data}.
However, training with GAN generated minority class samples can lead to boundary distortion~\cite{Santurkar2018Classification} resulting in performance drop for the majority class.
Mullick \textit{et al.},~\cite{Mullick2019Generative} proposed to generate minority class samples by convexly combining existing samples in an adversarial setting by fooling both the discriminator as well as the classifier.
Our proposed approach, on the other hand, generates additional minority samples from images belonging to the majority class.
In contrast to prior works that focus on only perturbing the whole image to generate a slightly different version of it, we use a translational generative model (Cycle-GAN) to perturb the \emph{content} (object of interest) without changing the \emph{context} much.
Our conjecture is: the classifier gets two different objects in the same context which would likely help the classifier to focus more on the object of interest rather than on the context for the disrciminative task.
Training classifiers with such image pairs where the two objects are in the same common context not only alleviates the dataset imbalance problem but also creates a more robust classifier by learning object features rather than focusing on the context. 

\section{Approach}
\label{sec:approach}

\begin{figure*}[!t]
  \centering
    \includegraphics[width=0.98\textwidth]{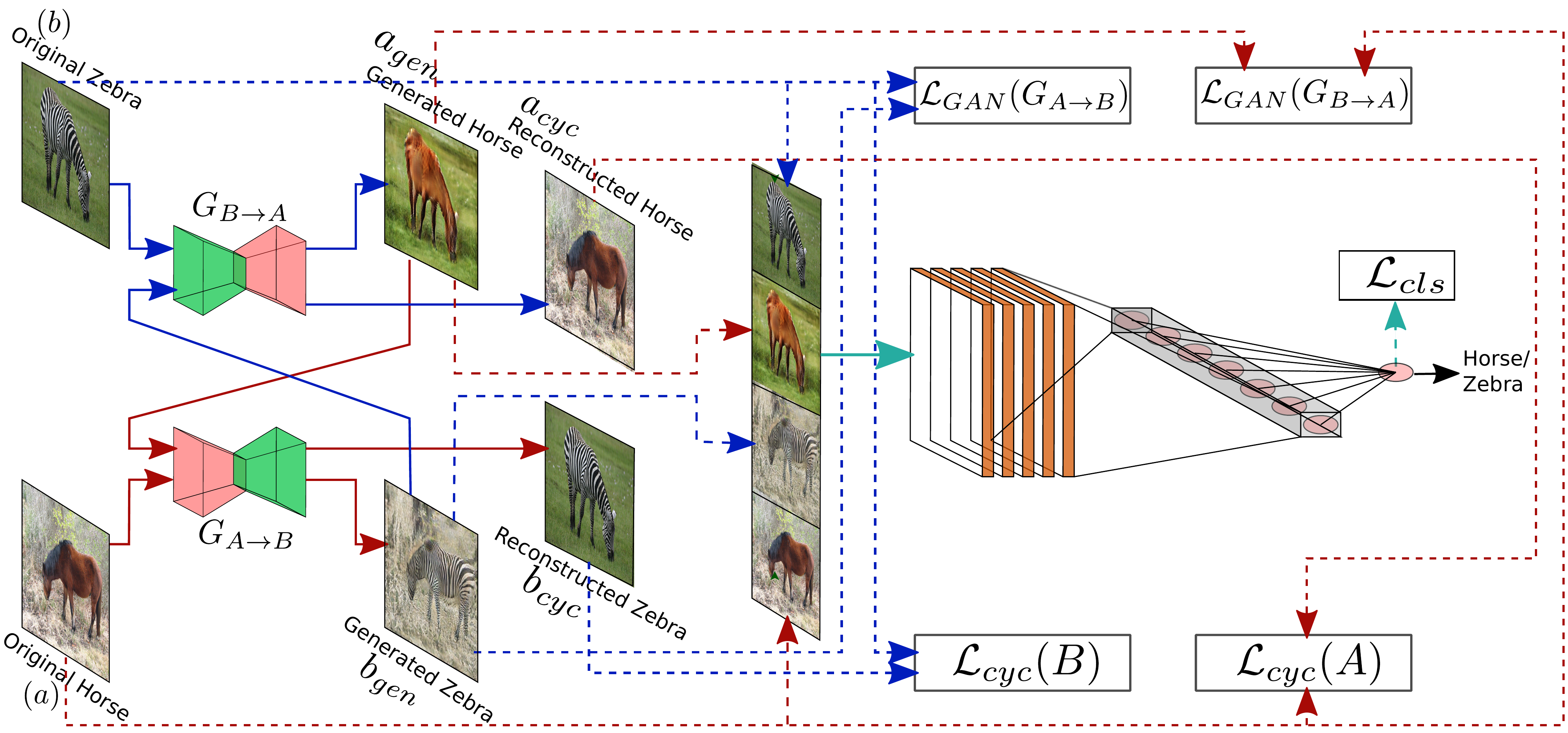}
  \setlength{\belowcaptionskip}{-6pt}
  \caption{\small \textbf{Approach overview.} We train a CNN to predict the presence of either horse or zebra in presence of high imbalance between the number of images of horses and zebras in the training data. In addition to images from the imbalanced dataset, the classifier uses images translated by cycle-GAN for training. What originally is a minority class becomes a majority class when translated. However, due to the obvious difference in quality between the generated and the real images they are weighed differently when used in training. The figure also shows the different losses along with the constituents contributing to them. Note that the identity loss for the cycle-GAN is not shown to make it less cluttered. The framework is trained by either keeping the GAN frozen throughout or alternatingly training the classifier and the GAN for a few epochs.}
  \label{fig:GAN_block_diag}
\end{figure*}

An overview of the training procedure is presented in Fig.~(\ref{fig:GAN_block_diag}).
The goal is to learn a generator to translate images from majority class to images of minority class so that the generated images make the dataset balanced for training.
We denote the majority class as $A$ and minority class as $B$.
In Fig.~(\ref{fig:GAN_block_diag}), horse images are assumed to be the majority class while the zebra images are assumed to be the minority class.

The proposed system uses a generator ($G_{A\rightarrow B}$) to translate a majority example ($a$) in the scene to an example ($b_{gen}$) belonging to the minority class.
Following the cycle-GAN philosophy, another generator ($G_{B\rightarrow A}$) will translate $b_{gen}$ to a majority class example ($a_{cyc}$).
Since $a_{cyc}$ is generated from the translated image $b_{gen}$ which, in turn, is generated from the original majority class example $a$, $a_{cyc}$ is known as the cyclic image of $a$.
The generated minority class image ($G_{A\rightarrow B}(a)$) along with a real minority class image ($b$) constitute the GAN loss ($\mathcal{L}_{GAN}$) while the real majority class image ($a$) and the cyclic image ($a_{cyc}$) will constitute the cycle-consistency loss ($\mathcal{L}_{cyc}$) as,

\begin{multline}
\mathcal{L}_{GAN}(G_{A\rightarrow B}, D_B, A, B) = -\mathbb{E}_{b\sim p_B(b)}\big[ \log D_B(b) \big]\\
\shoveright{- \mathbb{E}_{a\sim p_A(a)}\big[\log\big(1 - D_B\big( G_{A\rightarrow B}(a) \big)\big) \big]}\\
\shoveleft{\mathcal{L}_{cyc}(A) = \mathbb{E}_{a\sim p_A(a)}\big[ || G_{B\rightarrow A}\big( G_{A\rightarrow B} (a) \big) - a||_1 \big]}\\
\label{eq:a_to_b}
\end{multline}

where, $D_B$ denotes the discriminator.
Minimizing the GAN loss without the cycle-consistency constraint, can encourage the generator to map source domain images to a random permutation of the target domain images~\cite{Zhu2017Unpaired} as the mapped images still produce an identical target distribution.
This is, especially, true with imbalanced dataset where the network capacity can be too large with regard to the size of the dataset, facilitating the generator to memorize the target domain.
Cycle-consistency loss addresses this by putting further constraints on the generator by asking it to map back to the original domain if source-to-target and target-to-source generators are composed sequentially.
Following the original cycle-GAN formulation~\cite{Zhu2017Unpaired}, the cycle-consistency loss in this work, is measured as the $\ell_1$ distance between an image and its mapped back version.

A similar procedure provides the corresponding losses ($\mathcal{L}_{GAN}(G_{B\rightarrow A}, \allowbreak D_A, B, A)$ and $\mathcal{L}_{cyc}(B)$) for the scenario when we want to convert a visual scene containing majority class objects to a scene containing minority class objects.

\begin{multline}
\mathcal{L}_{GAN}(G_{B\rightarrow A}, D_A, B, A) = -\mathbb{E}_{a\sim p_A(a)}\big[ \log D_A(a) \big]\\
\shoveright{- \mathbb{E}_{b\sim p_B(b)}\big[\log\big(1 - D_A\big( G_{B\rightarrow A}(b) \big)\big) \big]}\\
\shoveleft{\mathcal{L}_{cyc}(B) = \mathbb{E}_{b\sim p_B(b)}\big[ || G_{A\rightarrow B}\big( G_{B\rightarrow A} (b) \big) - b||_1 \big]}\\
\label{eq:b_to_a}
\end{multline}
where, $D_A$ denotes the discriminator.

In addition, following~\cite{Taigman2016Unsupervised}, an identity loss ($\mathcal{L}_{ide}(G_{A\rightarrow B}, G_{B\rightarrow A})$) is used that encourages the generators to produce an identity image when a real sample of the target domain is passed through them.
\begin{multline}
\mathcal{L}_{ide}(G_{A\rightarrow B}, G_{B\rightarrow A}) = \mathbb{E}_{b\sim p_B(b)}\big[||G_{A\rightarrow B}(b) - b||_1\big] +\\
\mathbb{E}_{a\sim p_A(a)}\big[||G_{B\rightarrow A}(a) - a||_1\big]
\label{eq:identity_loss}
\end{multline}
Without this loss, translated images often come with additional tint.
This is, especially, problematic for the proposed framework as systematically tinted images at training time can confuse the classifier to predict almost randomly during inference where images are without such systematic tints.

The real (but abundant) and the translated (but originally rare) images will constitute a more balanced training set for the classifier $z:A\cup B\rightarrow [0,1]$, which provides a high value (close to 1) for a minority class image.
Unlike the traditional classifiers, our binary cross entropy loss function comprises of losses from four different types of images namely, - Real images from minority class ($b$), Generated samples of the minority class ($b_{gen}$) and the same corresponding to the majority class ($a$ and $a_{gen}$ respectively).
As the classifier provides the probability of an image to belong to the minority class, the loss coming from the minority class images is given by,
\small
\begin{gather}
\label{eq:cls_loss_min_class}
\hspace{-4mm}\mathcal{L}_{cls}^B\!=\!-\mathbb{E}_{b\sim p_B(b)}\big[\log z(b) \big]\!-\!\mathbb{E}_{a\sim p_A(a)}\big[\log z(G_{A\rightarrow B} (a))\big]
\end{gather}
\normalsize
Similarly the loss coming from the majority class images is given by,
\small
\begin{gather}
\hspace{-4mm}\mathcal{L}_{cls}^A\!=\!-\mathbb{E}_{a\sim p_A(a)}\big[\log(1- z(a)) \big]\!-\!\mathbb{E}_{b\sim p_B(b)}\big[\log(1-z(G_{B\rightarrow A} (b))) \big]
\label{eq:cls_loss_maj_class}
\end{gather}
\normalsize

The combined loss function for the classifier thus becomes,
\begin{equation}
\label{eq:classification_loss}
\mathcal{L}_{cls} = \mathcal{L}_{cls}^B + \frac{1}{\gamma}\mathcal{L}_{cls}^A
\end{equation}
where, $\gamma$ is the imbalance ratio that determines the relative weight of the loss components coming from the minority and the majority classes.
$\gamma$ is defined to be the ratio of the number of training examples in the majority class to the same in the minority class.
For a highly imbalanced scenario ($\gamma\!\!>>\!\!1$), the loss formulation penalizes the classifier much more when it misclassifies a minority class real image than when it misclassifies the same from the majority class.
This is to encourage the classifier to be free from the bias induced by the majority class.
The loss also penalizes the misclassification of images generated from the majority class more than the same generated from the minority class to help the classifier learn from the more diverse style captured by the GAN from the originally majority class images.

We propose two modes of training the proposed system - 1) \textbf{Augmented (AUG) Mode} 2) \textbf{Alternate (ALT) Mode}.
For both the modes, the cycle-GAN is first trained on the imbalanced dataset.
In augmented mode, the trained cycle-GAN is used throughout and acts as additional data generator for the classifier.
In this mode, only the classifier loss $\mathcal{L}_{cls}$ (ref. Eqn.~(\ref{eq:cls_loss_min_class})) is minimized.
In alternate mode, we alternately train either the cycle-GAN or the classifier for a few epochs keeping the other fixed.
The GAN is warm-started similar to the augmented mode of training while the classifier is trained from scratch.
During the training of the classifier only the classifier loss $\mathcal{L}_{cls}$ is backpropagated to change the classifier weights and the GAN acts as additional data generator similar to the augmented mode.
However, when the GAN is trained, the classifier acts as an additional teacher.
The full objective in this case is given by,
\small
\begin{multline}
\mathcal{L} = \mathcal{L}_{GAN}(G_{A\rightarrow B}, D_B, A, B) + \beta\mathcal{L}_{cyc}(A) +\\
\mathcal{L}_{GAN}(G_{B\rightarrow A}, D_A, B, A) + \beta\mathcal{L}_{cyc}(B) +\\
\alpha\mathcal{L}_{ide}(G_{A\rightarrow B}, G_{B\rightarrow A}) + \mathcal{L}_{cls}^A + \mathcal{L}_{cls}^B
\label{eq:overall_objective}
\end{multline}
\normalsize
where $\alpha$ and $\beta$ are hyperparameters weighing the relative importance of the individual loss terms.
We have also experimented with end-to-end training of the whole system which works good in a balanced data regime, but the performance is not satisfactory when there is high imbalance in data.
This may be due to the fact that the naturally less stable GAN training~\cite{Arjovsky2017Towards,Salimans2016Improved} gets aggravated in presence of high imbalance in data.
For example, with very few zebra and large number of horses, the cycle-GAN produces mostly the same image as the source horse with only a few faint stripes of zebra at different places of the image.
Getting good quality input at the beginning is critical for a convnet classifier as the deficits due to the ``noisy'' inputs at the beginning of the learning process of the classifier can not be overcome later~\cite{Achille2019Critical}.

\section{Experiments}
\label{sec:experiments}
The proposed approach is evaluated on three publicly available datasets namely CelebA~\cite{Liu2015Deep}, CUB-200-2011~\cite{WahCUB2011Caltech} and Horse2Zebra~\cite{Zhu2017Unpaired}.
Sections~\ref{susec:celeba}, \ref{susec:cub} and \ref{susec:horse2zebra} respectively provide the experimental details and evaluation results on them.

We evaluate the proposed approach in terms the improvement of performance of the classifier as well as that of the cycle-GAN in presence of high class imbalance in the training data.
The performance of the classifier is measured in terms of the $F_1$ score of the two classes so that the evaluation is fair irrespective of the skewness, if any, in the test data.
Following the common practice, we also report the classifier performance in terms of Average Class Specific Accuracy (ACSA)~\cite{Huang2016Learning,Mullick2019Generative}.
The best performing classifier on the validation set is used for reporting the test-set performances.
Due to space constraints, we provide only the $F_1$ score corresponding to the minority class in the main paper while the others are included in the \hyperref[appendix: Results]{\color{black}{appendix}}.
Although perceptual studies may be the gold standard for assessing GAN generated visual scenes, it requires fair amount of expert human effort.
In absence of it, inception score~\cite{Salimans2016Improved} is a good proxy.
However, for an image-to-image translation task as is done by a cycle-GAN, inception score may not be ideal~\cite{Barratt2018Note}.
This is especially true in low data regime as one of the requirements of evaluating using inception score is to evaluate using a large number of samples (\textit{i.e.}, 50k)~\cite{Salimans2016Improved}.
Thus, we have used an inception accuracy which measures how good an inception-v3~\cite{Szegedy2016Rethinking} model trained on real images can predict the true label of the generated samples.
This is given by the accuracy of the inception-v3 model on the images translated by the cycle-GANs.
A cycle-GAN providing higher accuracy to the same set of test images after translation, is better than one giving a lower accuracy.
For this purpose, the ImageNet pretrained inception-v3 model is taken and the classification layer is replaced with a layer with two neurons.
This modified network is then finetuned on balanced training data that is disjoint from the test set for all the three datasets.
For each dataset the evaluation procedure is repeated 5 times while keeping the data fixed but using independent random initialization during runs. The average result of these 5 runs is reported.

\subsection{Experiments on CelebA}
\label{susec:celeba}
Many datasets have contributed to the development and evaluation of GANs.
However, the CelebA~\cite{Liu2015Deep} dataset has been the most popular and canonical in this journey.
CelebA contains faces of over 10,000 celebrities each of which has 20 images.
We used images from this dataset to study the effect of imbalance for a binary classifier predicting the gender from the face images only.
Gender bias due to imbalance in datasets~\cite{Miltenburg2016Stereotyping,Hendricks2018Women} has started to gain attention recently and face images make the problem more challenging as many attributes can be common between males and females or can only be very finely discriminated.

\noindent{\textbf{Experimental Setup}}: We took a subset of 900 female images and considered this to be the majority class.
The male images are varied in number from 100 to 900 in regular intervals of 100.
In addition, we experimented with 50 male images too which makes the highest imbalance ratio ($\gamma$) for this dataset to be 18.
We follow the same architecture of the cycle-GAN as that of Zhu \textit{et al.}~\cite{Zhu2017Unpaired} where the generator consists of two stride-2 convolutions, 9 residual blocks, and 2 fractionally-strided convolutions.
For discriminator, similarly, a 5 layer CNN is used for extracting features, following the PatchGAN~\cite{Isola2017Image} architecture.
To avoid over-fitting in presence of less data, our binary classifier is a simple CNN without any bells and whistles.
We used a classifier with same feature extraction architecture as that of the discriminator used above.
The initial convolution layers are followed by two fully-connected layers of 1024 and 256 neurons along with a output layer consisting of two neurons.
To handle over-fitting further, we used dropout with probability $0.5$ of dropping out neurons from the last fully-connected layer.
Following~\cite{Zhu2017Unpaired}, we have used $\alpha = 5$ and $\beta = 10$ as the loss weights for the cycle-GAN throughout the experiments.
The test set for this dataset comprises of 300 images of male and female each.

\begin{table*}[t]
\centering
\resizebox{\linewidth}{!}{ 
\begin{tabular}{l||cccccc||cccccc}
\hline
\textbf{Dataset} & \multicolumn{6}{c}{\textbf{CelebA}} & \multicolumn{5}{c}{\textbf{CUB-200-2011}}\\
\hline
\#Minority examples              & 50              & 100             & 200             & 300             & 400             & 500     &12         & 25		  & 50 		& 75  & 100 & 125\\
\hhline{=#======#======}
Vanilla                      & 0.1500          & 0.5220          & 0.7740          & 0.8460          & 0.9020          & 0.9160      &0.0240      & 0.1180 		& 0.2960 & 0.3700 & 0.5520 & 0.6160 \\
\hline
TS                 & 0.1560          & 0.6300          & 0.7880          & 0.8420          & 0.8960          & 0.9200   &0.0240 & 0.1180 &0.2880 & 0.3640 & 0.2567 & 0.6080 \\
CS               & 0.7825          & 0.8012          & 0.8975          & 0.9137          & 0.9250          & 0.9244    &0.3674        & 0.5007 		& 0.5001 & 0.6384 & 0.6485 & 0.7212 \\
US                & 0.8029          & 0.8491          & 0.9036          & \textbf{0.9176} & 0.9179          & \textbf{0.9307}  &0.2952 & 0.4263 		& 0.6074 & 0.6893 & \textbf{0.7169} & 0.7080 \\
OS                 & 0.5805          & 0.7333          & 0.8749          & 0.9036          & 0.9181          & 0.9188  &0.0602         & 0.1943 		& 0.3910 & 0.5007 & 0.6295 & 0.6322 \\
US + CS & 0.8041          & 0.8463          & 0.9019          & 0.9163          & 0.9191          & 0.9220    &0.5394       & 0.4760 		& 0.6655 	& 0.6647 & 0.6995 & 0.7201 \\
OS + CS  & 0.7644          & 0.8249          & 0.8916          & 0.9065          & 0.9223          & 0.9260   &0.3845        & 0.4225		& 0.5739 & 0.6147 & 0.6246 & 0.6913 \\
SMOTE~\cite{Chawla2002Smote}                        & 0.6208          & 0.7685          & 0.8807          & 0.8895          & 0.9167          & 0.9208   &0.0586        & 0.4533 		& 0.6375 & 0.5625 & 0.6708 & 0.6674 \\
CBL~\cite{Cui2019Class} ($\beta\!=\!0.9$)                 & 0.6736          & 0.7771          & 0.8867          & 0.9061          & 0.9118          & 0.9206    &0.1342       & 0.4006 		& 0.5068 & 0.5624 & 0.6187 & 0.6890 \\
CBL~\cite{Cui2019Class} ($\beta\!=\!0.99$)               & 0.7012          & 0.8021          & 0.8938          & 0.9118          & 0.9178          & 0.9226      &0.3259     & 0.5392 		& 0.6001 & 0.6196 & 0.6369 & 0.6600 \\
CBL~\cite{Cui2019Class} ($\beta\!=\!0.999$)              & 0.7692          & 0.8250          & 0.8922          & 0.9122          & 0.9179          & 0.9220   &0.3492        & 0.5256 		& 0.6105& 0.5400 & 0.5937 & 0.7212 \\
CBL~\cite{Cui2019Class} ($\beta\!=\!0.9999$)              & 0.7885          & 0.8099          & 0.8977          & 0.9127          & 0.9241          & 0.9226    &0.3562       & 0.5950 		& 0.5138 & 0.5933 &  0.6547 & \textbf{0.7344} \\
\hline
(ours) ALT Mode              & \textbf{0.8240} & 0.8520          & 0.8900          & 0.8880          & 0.8520          & 0.8920     &0.5120        & 0.5120 		& 0.5640 & 0.6340 & 0.6940 &0.5960 \\
(ours) AUG Mode               & 0.8060          & \textbf{0.8740} & \textbf{0.9140} & 0.9160 & \textbf{0.9340} & 0.9220   &\textbf{0.5940}         & \textbf{0.6040} 		& 		\textbf{0.6680} & \textbf{0.7060} & 0.7040 & 0.7180 \\
\hline
\end{tabular}}
\vspace{2mm}
\caption {Comparison of the proposed method on the CelebA and CUB-200-2011 datasets using both AUG and ALT mode. The left half of the table shows the results for CelebA and the right half shows the results for the pairs of birds chosen from CUB-200-2011 dataset. The number of majority class training images (female) in CelebA is fixed to 900 while the CUB-200-2011 majority class (Sparrow) images 250. Minority class $F_1$ score for both are shown where the individual column headings indicate the number of minority class training images.}
\label{tab:celeba_cub}
\end{table*}

\noindent{\textbf{Results}}: Table~\ref{tab:celeba_cub} shows a comparative evaluation of the classifier performance of our proposed model for both AUG and ALT mode.
In this table, we have provided the results for upto 500 male images due to space constraint and the rest are provided in the \hyperref[appendix: Results]{\color{black}{appendix}}. 
We compared with some of the classic methods as well as methods particularly used with deep learning based classifiers.
Namely they are random oversampling of minority class (OS), random undersampling of majority class (US), class frequency based cost sensitive weighing of loss (CS)~\cite{Lawrence1998Neural}, adjusting decision threshold by class frequency at prediction time (TS) and SMOTE~\cite{Chawla2002Smote} are some of the classical methods with which we compare our proposed method.
We have also studied the effect of combining a few of the above classical methods and are reporting the two combinations - undersampling with cost sensitive learning (US+CS) and oversampling with cost sensitive learning (OS+CS).
We have also compared with recent state-of-the-art approach using a class balanced loss (CBL)~\cite{Cui2019Class} for deep learning based classifiers.

We started with a highly imbalanced dataset of male and female faces where we randomly took 50 male and 900 female face images.
Training the classifier without any data balancing strategy (we refer this classifier as the `vanilla' classifier) misclassifies almost all of the minority class examples (male) as majority class examples (female) resulting in a accuracy of only 0.08 for males while the accuracy for the female images is perfect (\textit{i.e.}, 1).
The $F_1$ score of the minority class for this case is 0.15.
Our AUG mode of training improves the accuracy more than 5 fold to 0.8060 while the ALT mode of training improves it further to 0.8240.
The fact that the classification is more balanced after the proposed dataset repairment is corroborated by the simultaneous increase of the precision of the majority class (female) from 0.5200 to 0.7720 (AUG mode) and 0.8260 (ALT mode).
The male classification accuracy in AUG and ALT mode reaches up to 0.736 and 0.820 respectively.
Table~\ref{tab:celeba_cub} shows the minority class $F_1$ score comparison with the state-of-the-arts.
For 4 out of the 6 highly skewed regions (imbalance ratio ranging from 18 to 1.8) our proposed method significantly outperforms the rest.
Out of the two cases, our augmentation mode is a close second (for 900:300 scenario).
As the imbalance increases, our relative performance is better and better compared to other approaches.
Performing better in highly skewed data distribution is very important as high imbalance implies more difficult case.
This shows that our approach can deal with the hard cases in the challenging CelebA dataset.

\begin{figure*}[!t]
\centering
\subfigure[]{
\label{fig:celeba_50_900_bias}
\includegraphics[width=0.235\linewidth]{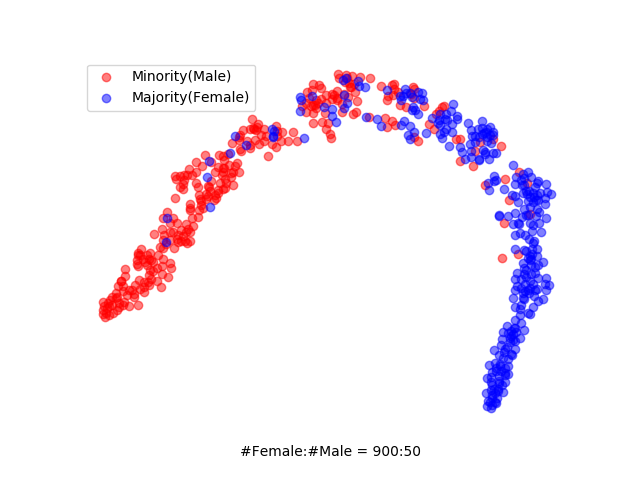}}
\subfigure[]{
\label{fig:celeba_50_900_aug}
\includegraphics[width=0.235\linewidth]{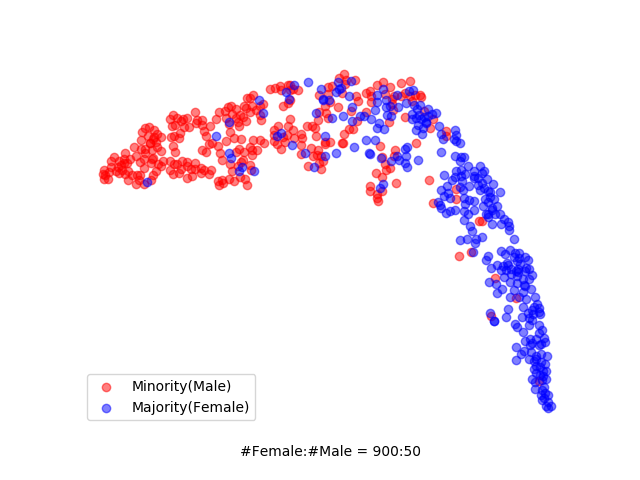}}
\subfigure[]{
\label{fig:celeba_100_900_bias}
\includegraphics[width=0.235\linewidth]{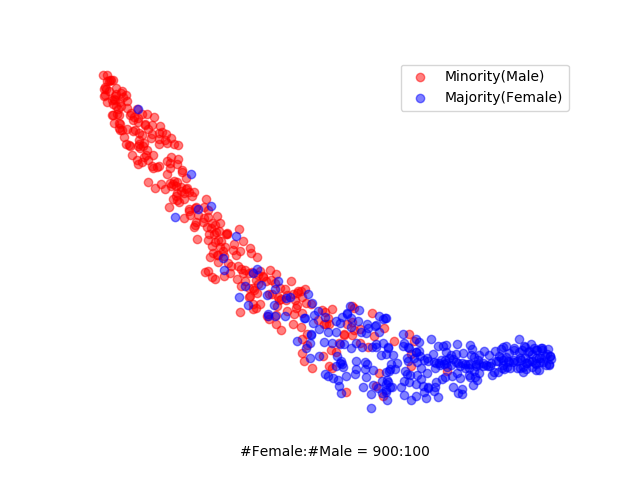}}
\subfigure[]{
\label{fig:celeba_100_900_aug}
\includegraphics[width=0.235\linewidth]{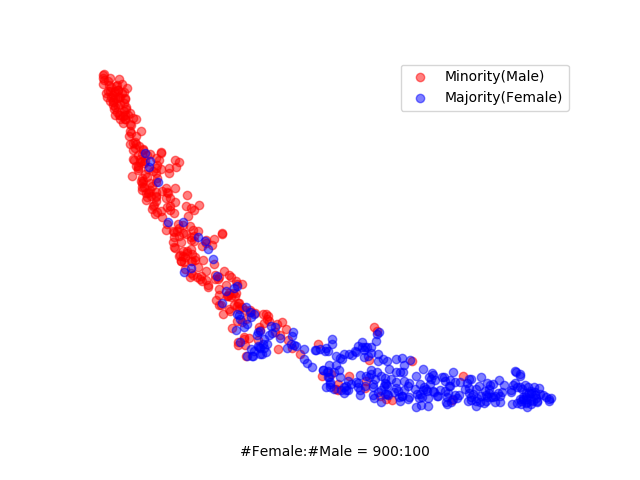}}
\subfigure[]{
\label{fig:Z2H_25_450_bias}
\includegraphics[width=0.235\linewidth]{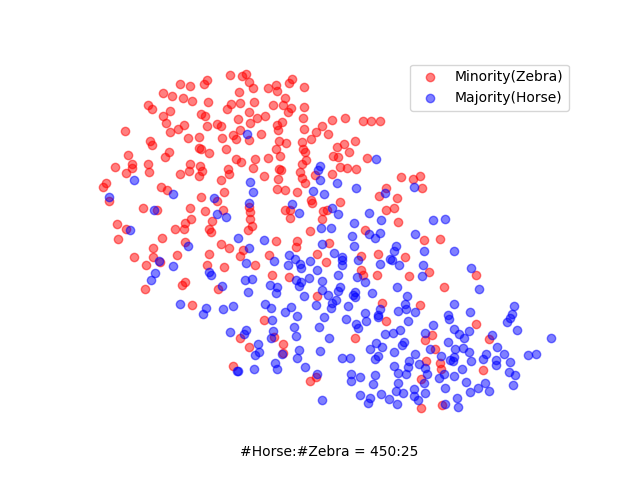}}
\subfigure[]{
\label{fig:Z2H_25_450_aug}
\includegraphics[width=0.235\linewidth]{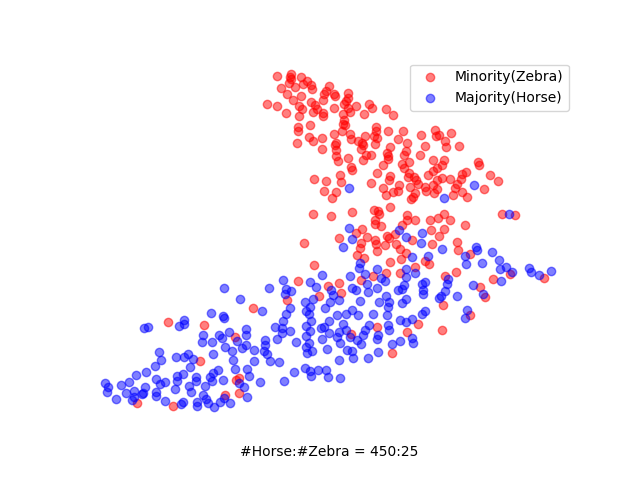}}
\subfigure[]{
\label{fig:z2h_50_450_bias}
\includegraphics[width=0.235\linewidth]{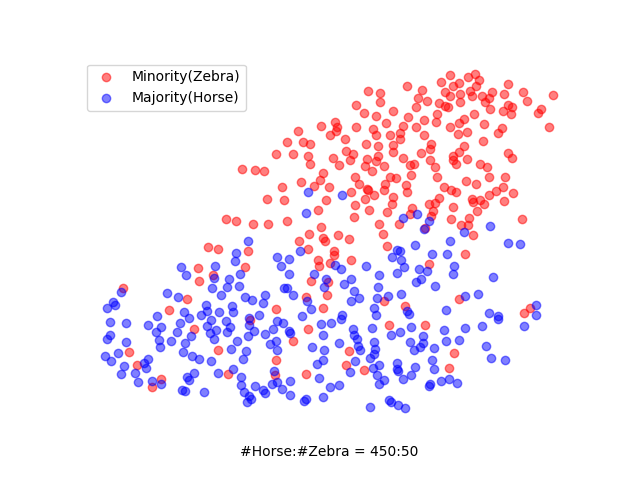}}
\subfigure[]{
\label{fig:z2h_50_450_aug}
\includegraphics[width=0.235\linewidth]{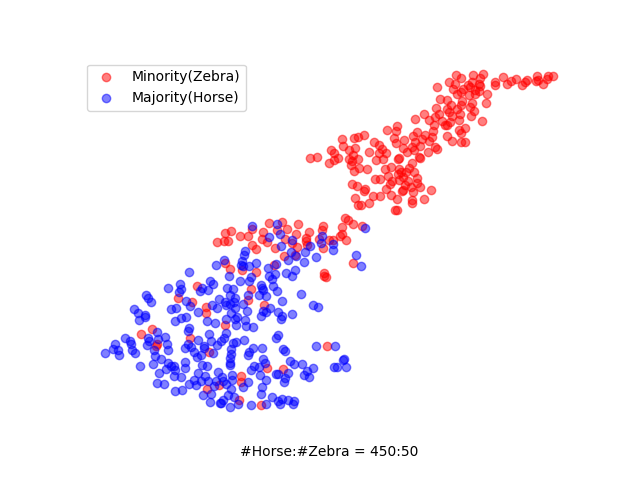}}
\caption{t-SNE visualization for CelebA and Horse2Zebra dataset. Top row contains the t-SNE visualizations for 2 different imbalance ratio on CelebA dataset. Blue color indicates representation of majority class which is female while red color indicates the representation of the minority class i.e male. Fig.~\ref{fig:celeba_50_900_bias} and  \subref{fig:celeba_50_900_aug} show the visualisation from the imbalanced (vanilla) model and augmented model for 900:50 imbalance ratio respectively. Fig.~\ref{fig:celeba_100_900_bias} and \subref{fig:celeba_100_900_aug} represent the 900:100 imbalance ratio. Similarly in the bottom row, Fig.~\ref{fig:Z2H_25_450_bias} and \subref{fig:Z2H_25_450_aug} show the visualizations for imbalance ratio of 450:25 on Horse2Zebra dataset and Fig.~\ref{fig:z2h_50_450_bias} and \subref{fig:z2h_50_450_aug} show the results for imbalance ratio of 450:50 respectively. Here, blue color indicates Horses (majority class) while the red color indicates Zebras (minority class). As can be seen from all the figures, a clear separation of the classes can be visualized . Best viewed in color.}
\label{fig:tsne}
\end{figure*}

We have also provided t-SNE~\cite{Maaten2008Visualizing} visualizations of the learned feature representation from the last fully-connected layer the classifier before and after the dataset repairment.
Fig.~\ref{fig:celeba_50_900_bias} and \subref{fig:celeba_50_900_aug} show the visualization of the test set features for the vanilla classifier and the same trained in augmented mode respectively.
This is for the case when the number of minority training examples is 50 while the majority training example is 900.
It can be seen that the separation between the two classes is better after the repairment even in presence of such high imbalance.
The same can be seen for the imbalanced training with 100 male and 900 female faces as shown in Fig.~\ref{fig:celeba_100_900_bias} (vanilla) and \subref{fig:celeba_100_900_aug} (AUG).

Though in the high imbalance region (towards left of Table~\ref{tab:celeba_cub}), both the modes of our proposed approach perform very good, the augmentation mode of training is consistently better than the alternate mode of training.
The significance of the alternate mode of training, though, lies in the fact that it allows the cycle-GAN to improve itself by helping the classifier to discriminate better in presence of highly skewed training data.
The improvement of performance of the GAN in terms of inception accuracy is shown in Table~\ref{tab:gan_perf}.
Except for one scenario, the GAN performance improves for all the rest.
Thus, the proposed approach provides a choice between going for bettering the GAN by trading off a little in the classifier performance (ALT mode) or going for improving the classifier without changing the GAN (AUG mode).

\begin{table*}[t]
\centering
\begin{adjustbox}{width=\textwidth}
\begin{tabular}{c|cc||c|cc||c|cc}                          
\multicolumn{3}{c}{\textbf{Celeb A Dataset}}                      & \multicolumn{3}{c}{\textbf{CUB-200-2011 Dataset}}            & \multicolumn{3}{c}{\textbf{Horse2Zebra Dataset}}          \\
\hline
\# Male & CycleGAN        & ALT Mode & \#Flycatcher & CycleGAN & ALT Mode & \#Zebra & CycleGAN & ALT Mode \\
\hline
50                & \textbf{0.4798} & 0.4677                      & 12                    & 0.4149          & \textbf{0.4554}                            &                         &         &                             \\
100               & 0.5574          & \textbf{0.6220}             & 25                   &  \textbf{0.4238}        &  0.4059                           &  25                       & \textbf{0.3532}         & 0.3480                    \\
200               & 0.7646          & \textbf{0.8152}             & 50                   &  0.4337        &  \textbf{0.4505}                           & 50                        & \textbf{0.3880}         & 0.2680                            \\
300               & 0.7516          & \textbf{0.8200}             &  75                  &  0.4347        &  \textbf{0.4653}                          &75                        & \textbf{0.5000}         & 0.4440                           \\
400               & 0.8122 & \textbf{0.8455}                      &  100                  &  \textbf{0.4317}        &  0.4307                           &  100                       & \textbf{0.5492}         &   0.4780                           \\
500               & 0.8286          & \textbf{0.8457}             &  125                  &  0.4198        & \textbf{0.4604}                            &                         &          &                             \\
                       
\hline
\end{tabular}
\end{adjustbox}
\bigskip
\caption {Comparison of the inception accuracies of the Cycle-GAN before and after applying our proposed dataset repairment approach (in ALT mode). The table is divided into 3 parts from left to right corresponding to the three datasets on which the experimentations are performed. The dataset names are marked as headings of the parts. Each part is further divided into two halves where the left half shows the performance for the vanilla cycle-GAN and the right half shows the same after the cycle-GAN is trained in alternate mode of the proposed approach. The values shown are averaged over 5 runs for each of the cases. For two (CelebA and CUB-200-2011) out of the three datasets, the ALT mode of training improves the Cycle-GAN performance for most of the cases, while for Horse2Zebra dataset the performance is not increasing which may be due to the distintive feature of the dataset where the attributes of the two classes are markedly different.}
\label{tab:gan_perf}
\end{table*}

\begin{figure}[t]
	\begin{center}
		\includegraphics[width=0.95\linewidth]{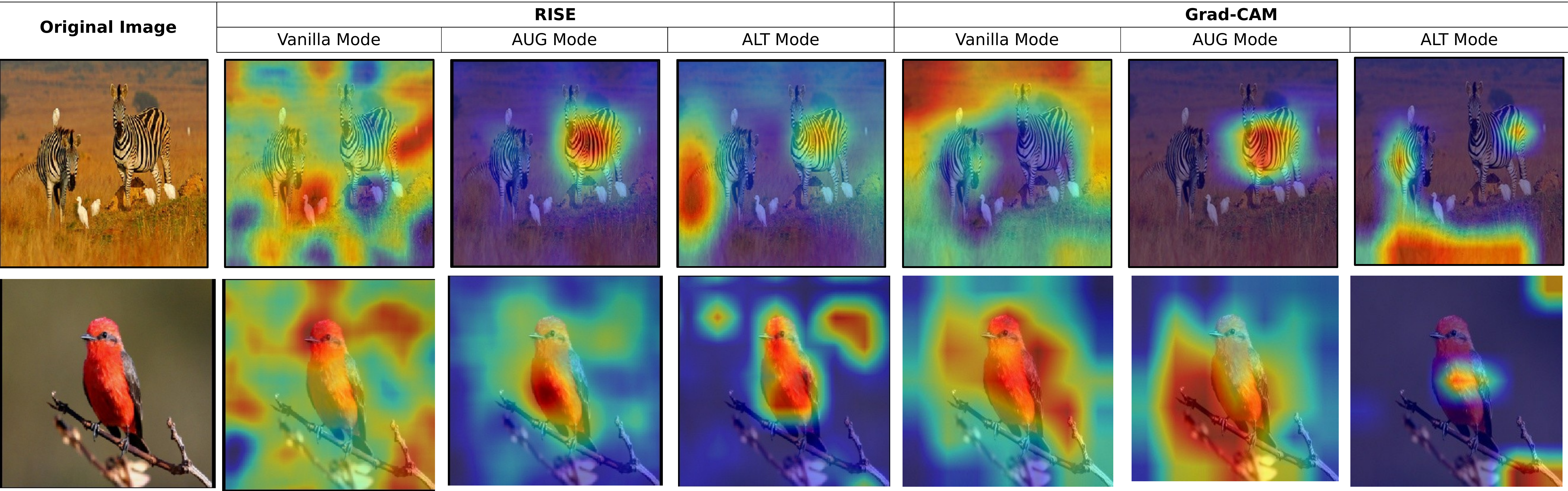}
	\end{center}
	\caption{\small Explanation heatmaps of images from Horse2Zebra (first row) and CUB (second row). Leftmost column shows the images while col 2-4 and 5-7 show RISE and Grad-CAM heatmaps respectively (importance increases from blue to red). The column sub-headings denote the corresponding classifier. Note that the heatmaps are generated for the predicted output by the classifiers. For vanilla mode, the prediction is wrong (horse and sparrow respectively) while for both AUG and ALT mode they are right. An interesting observation is revealed by the RISE generated heatmap (first row $4^{th}$ col) showing that the evidence for ALT mode classifier is coming from a third zebra hidden in the grass. Best viewed in color.}
	\label{fig:saliency}
	\vskip -0.2in
\end{figure}

\noindent\textbf{Visualizing important attributes for classification}:
We aim to visually confirm that the proposed approach helps the classifier to look at the right evidences.
For this purpose, we rely on two visual explanation techniques: Grad-CAM~\cite{Selvaraju2017Gradcam} and RISE~\cite{Petsiuk2018Rise}.
They provide explanation heatmaps pointing to locations in the input image that are most important for a decision.
Fig.~\ref{fig:saliency} shows visualizations for some representative minority images when they are passed through the vanilla classifier and the same after training in AUG and ALT mode.
The top row shows a representative zebra image from the Horse2Zebra dataset. The vanilla model tends to concentrate on the surroundings (context) to classify (incorrectly), while the proposed models accurately concentrate on the zebras (object) for correct classification.
Similarly, the bottom row shows a flycatcher image from CUB dataset. When classifying the image, the vanilla model tends to scatter its attention while the proposed models base their prediction on right areas
by learning to ignore inappropriate evidences.

\subsection{Experiments on CUB}
\label{susec:cub}
CUB-200-2011 birds dataset~\cite{WahCUB2011Caltech} is a popular benchmark dataset for fine-grained image classification.
It contains 11,788 images of 200 different classes of North American bird species.
Fine-grained classification is inherently difficult task as the members of the classes have very similar attributes with small inter-class variation requiring the classifier to focus on finer aspects of the images that are both descriptive and adequately discriminative.
The challenges of fine-grained classification requires a lot of parameters to learn~\cite{Gao2016Compact,Lin2015Bilinear,Zhang2019Learning} which in turn needs a lot of training examples for proper training.
Thus fine-grained classification provides us with an interesting testcase for low-data high-bias regime of classification.

\noindent{\textbf{Experimental Setup}}: For the experimentations, we have chosen two similar looking bird species (Flycatcher and Sparrow).
We took all the images (226) of flycatcher that the dataset has and considered it to be the minority class.
The 226 flycatcher images are split into 125 train, 50 val and 51 test images.
The number of majority class (sparrow) images were kept constant at 250.
We vary the number of flycatcher images from 25 to 125 in steps of 25 making the imbalance ratio vary from 10 to 5 in steps of 1.
In addition, we also experimented with only 12 flycatcher images.
The val and test set size of sparrow is kept same as those of the flycatcher class.
The model architectures are kept same as those used for CelebA experiments (ref. Sec~\ref{susec:celeba}).

\noindent{\textbf{Results}}: Comparative evaluations of the proposed approach for the pair of birds in the CUB-200-2011 data set is shown in the rightmost part of Table~\ref{tab:celeba_cub}.
The table provides the $F_1$ score of the minority class for the different approaches.
It can be observed that our proposed method in augmentation mode is outperforming other approaches for highly skewed training data.
The alternate mode of training the classifier does not perform as good as it does for the face images.
However, the cycle-GAN inception accuracy gets improved for 4 out of 6 cases as can be seen in the middle part of Table~\ref{tab:gan_perf} where for highly skewed training data, the alternate model betters the vanilla cycle-GAN.

\subsection{Experiments on Horse2Zebra}
\label{susec:horse2zebra}

Cycle-GAN~\cite{Zhu2017Unpaired} has been particularly successful in translating between images of horses and zebras.
Unlike finegrained bird images, horses and zebras have distinctive features with marked difference in skin texture which allows a classifier to perform well even with very less data.
To test with this end of the spectrum, our proposed approach was applied for a classification task between horse and zebra images in presence of high imbalance in training data.

\noindent{\textbf{Experimental Setup}}: The Horse2Zebra dataset~\cite{Zhu2017Unpaired} contains 1187 images of horses and 1474 images of zebras.
We took the horse class as majority and used a total of 450 Horse images.
The number of minority class (zebra) examples were varied from 25 to 100 in steps of 25 allowing us to experiment with low data and low balance regime.
The val and test set consists of 100 and 150 images respectively for each of the animal categories.
For this case also, the model architectures are kept same as those used for CelebA experiments (ref. Sec~\ref{susec:celeba}).

\begin{table}[t]
\centering

\resizebox{1.02\linewidth}{!}{  
\begin{tabular}{l||c|c|c|c|c|c|c|c|c|c|c|c||c|c}
\hline
\multirow{2}{*}{\textbf{\#Zebra}} & \multirow{2}{*}{Vanilla} & \multirow{2}{*}{TS} & \multirow{2}{*}{CS} & \multirow{2}{*}{US} & \multirow{2}{*}{OS} & \multirow{2}{*}{US+CS} & \multirow{2}{*}{OS+CS} & \multirow{2}{*}{SMOTE~\cite{Chawla2002Smote}} & \multicolumn{4}{c||}{CBL~\cite{Cui2019Class}} & \multicolumn{2}{c}{Ours}
\\
&&&&&&&&& $\beta=0.9$ & $\beta=0.99$ & $\beta=0.999$ & $\beta=0.9999$ & ALT & AUG\\
\hhline{=#=|=|=|=|=|=|=|=|=|=|=|=#=|=}
25& 0.0500 & 0.0340 & 0.7925 & 0.1333 & 0.4750 & 02865 & 0.7366 & 0.2108 & 0.7613 & 0.8116 & 0.8263 & 0.8120 & 0.8040 & \textbf{0.8500}\\
50 & 0.5580 & 0.6260 & 0.8329 & 0.7891 & 0.8012 & 0.7343 & 0.8522 & 0.7298 & 0.8347 & 0.8316 & 0.8222 & 0.8192 & 0.8320 & \textbf{0.8620}\\
75 & 0.7180 & 0.7040 & 0.8699 & 0.8457 & 0.8370 & 0.8644 & 0.8680 & 0.7902 & 0.8562 & 0.8587 & 0.8524 & 0.8633 & 0.8020 & \textbf{0.8780}\\
100 & 0.7680 & 0.7940 & 0.8684 & 0.8521 & 0.8628 & \textbf{0.8711} & 0.8735 & 0.8254 & 0.8446 & 0.8595 & 0.8683 & 0.8657 & 0.8220 & 0.8520\\
\hline
\end{tabular}}
\vspace{0.1mm}
\caption {\small Performance comparison on Horse2Zebra dataset.
The first row lists the approaches against which we have compared our performance.
The number of majority class training images (horses)
is fixed at 450 while the number of zebra images used in training is shown as the row headings.
The minority class $F_1$ score is seen to get improved with application of our proposed approach especially in the augmented mode for high imbalance regions.
}
\label{tab:horse2zebra}
\vskip -0.2in
\end{table}

\noindent{\textbf{Results}}: Table~\ref{tab:horse2zebra} shows comparative evaluations of the proposed approach for this dataset in terms of the $F_1$ score of the minority class (zebra).
The proposed method in AUG mode of training comprehensively outperforms the other approaches in the high imbalance region with the difference in $F_1$ score with the second best approach reaching almost 2.5\% for the highest imbalance ratio with the number of training images for zebra being only 25.
The inception accuracies of the cycle-GAN before and after training with ALT mode is provided in the rightmost part of Table.~\ref{tab:gan_perf}.
For this dataset we got consistently worse GANs while training in this mode.
This can be due to the distinctive texture between the two animals that aids classification but imposes difficulty in translation especially in presence of the classifier loss coming from noisy data.

\section{Conclusion}
\label{sec:conclusion}

We propose a joint generation and classification network to handle severe data imbalance by transforming majority class examples to minority class and vice versa. Our approach explores a Cycle-consistent GANs that not only generates training samples that are much broader and diverse but also balanced in terms of number of examples. Experiments on three standard datasets demonstrate that both classifier and the GAN achieves superior performance than existing data imbalance mitigation approaches.

\noindent{{\textbf{Acknowledgements:}} This work was partially supported by the IIT Kharagpur ISIRD program and the SERB Grant SRG/2019/001205.}

{\small
\bibliographystyle{splncs04}
\bibliography{egbib}
}

\pagebreak
\appendix
\addcontentsline{toc}{section}{Appendix}
\section*{Appendix}
\label{appendix: Results}
\section{Training Details}

\subsection{On CelebA Dataset}
\label{appendix: celeb}
\begin{itemize}
    \itemsep0em
    \item The Cycle-GAN was trained from scratch on the imbalance data with a learning rate of 2e-4, and weights initialized from a Gaussian distribution $N(0, 0.02)$. We follow~\cite{Zhu2017Unpaired} for the hyperparameters and choice of optimizers. The learning rate was kept constant through out the training. The image size used for training is 256x256.
    \item We use the 200-Epoch trained models for generating the translated images and using them for training the classifier in \textbf{AUG-Mode}.
    \begin{itemize}
        \itemsep0em
        \item In \textbf{AUG-Mode}, we train the Classifier for 20 Epochs and use a validation-set containing 300 images from each class to select the best model.
    \end{itemize}
    \item For \textbf{ALT-Mode} training, we use the weights of the 200-Epoch models of Vanilla Cycle-GAN to warm-start the Cycle-GAN part of the ALT-Mode Model and the Classifier part is initialized with weights taken from a Gaussian distribution of $N(0, 0.02)$. 
    \begin{itemize}
        \itemsep0em
        \item The interval for swapping the training in AUG-Mode was kept to be 5-Epochs.
        \item The Classifier is trained for the first 5-Epochs keeping the Cycle-GAN part frozen.
        \item In the next 5-Epochs, the Cycle-GAN is trained keeping the Classifier part frozen, but in this case the additional loss obtained from the Classifier part is also back-propagated through the Cycle-GAN part.
        \item The above mentioned training-swap procedure is followed till the models are trained for a total of 100-Epochs.
    \end{itemize}
    \item For evaluating the GANs we take the GAN models obtained after the 100-Epoch training in ALT-Mode and compare them with the corresponding models obtained after 300-Epoch training of Vanilla Cycle-GAN. The 300-Epoch models of Vanilla Cycle-GAN were chosen to provide a fair comparison of the GAN Performance as in the AUG-Mode, we warm-start the Cycle-GAN part with 200-Epoch Models. 
\end{itemize}


\subsection{On CUB Dataset}
\label{appendix: CUB}
\begin{itemize}
    \itemsep0em
    \item The Cycle-GANs were trained in the same fashion as used in case of the CelebA Dataset with some changes mentioned below. The learning rate was kept constant for the first 50 Epochs and is linearly decayed to zero over the next 150 Epochs of training.
    \item For CUB, we use the 50-Epoch trained models for \textbf{AUG-Mode}.
    \begin{itemize}
        \itemsep0em
        \item In \textbf{AUG-Mode}, all procedures are kept same as that of CelebA but we use a validation-set containing 50 images from each class to select the best model.
    \end{itemize}
    \item For \textbf{ALT-Mode} training, we use the weights of the 50-Epoch models of Vanilla Cycle-GAN to warm-start the Cycle-GAN part of the ALT-Mode Model and the rest are same as that in case of CelebA Dataset. 
    \begin{itemize}
        \itemsep0em
        \item The training-swap procedure is followed till the models are trained for a total of 50-Epochs.
    \end{itemize}
    \item For evaluating the GANs we take the GAN models obtained after the 50-Epoch training in ALT-Mode and compare them with the corresponding models obtained after 100-Epoch training of Vanilla Cycle-GAN. 
\end{itemize}

\subsection{On Horse2Zebra Dataset}
\label{appendix: H2Z}
\begin{itemize}
    \itemsep0em
    \item All the training procedures and the Epoch-checkpoints for using the models are kept same as that in the case of the CUB Dataset, except we use a validation-set containing 100 images from each class in this case. 
\end{itemize}

\begin{table*}[htbp]
\centering
\resizebox{\linewidth}{!}{ 
\begin{tabular}{l||cccccc||cccccc}
\hline
\textbf{Dataset} & \multicolumn{8}{c}{\textbf{CelebA}}\\
\hline
\#Minority examples              & 50              & 100             & 200             & 300             & 400             & 500  &600 &700 &800 &900  \\
\hhline{=#======#======}
Vanilla                      & 0.1500          & 0.5220          & 0.7740          & 0.8460          & 0.9020          & 0.9160  &0.9260 &0.9320 &0.9380 &\textbf{0.9400}        \\
\hline
US                & 0.8029          & 0.8491          & 0.9036          & \textbf{0.9176} & 0.9179          & \textbf{0.9307} & 0.9223 & 0.9152 & 0.9216 & 0.9236 \\
TS                 & 0.1560          & 0.6300          & 0.7880          & 0.8420          & 0.8960          & 0.9200   & 0.9260  &0.9320 &0.9380 &\textbf{0.9400}  \\
CS               & 0.7825          & 0.8012          & 0.8975          & 0.9137          & 0.9250          & 0.9244  & 0.9250 & 0.9267 & 0.9256 & 0.9359   \\

OS                 & 0.5805          & 0.7333          & 0.8749          & 0.9036          & 0.9181          & 0.9188   & 0.9215 & 0.9286 & 0.9321 & 0.9329 \\
US + CS & 0.8041          & 0.8463          & 0.9019          & 0.9163          & 0.9191          & 0.9220  & 0.9259 & 0.9277 & 0.9321 & 0.9250   \\
OS + CS  & 0.7644          & 0.8249          & 0.8916          & 0.9065          & 0.9223          & 0.9260   & 0.9310 & 0.9313 & 0.9297 & 0.9393 \\
SMOTE~\cite{Chawla2002Smote}                        & 0.6208          & 0.7685          & 0.8807          & 0.8895          & 0.9167          & 0.9208 & \textbf{0.9285} & 0.9350 & \textbf{0.9387} & 0.9357 \\
CBL~\cite{Cui2019Class} ($\beta\!=\!0.9$)                 & 0.6736          & 0.7771          & 0.8867          & 0.9061          & 0.9118          & 0.9206   & 0.9260 &\textbf{0.9376} & 0.9282 & 0.9384 \\
CBL~\cite{Cui2019Class} ($\beta\!=\!0.99$)               & 0.7012          & 0.8021          & 0.8938          & 0.9118          & 0.9178          & 0.9226   & 0.9190 & 0.9282 & 0.9348 & 0.9380   \\
CBL~\cite{Cui2019Class} ($\beta\!=\!0.999$)              & 0.7692          & 0.8250          & 0.8922          & 0.9122          & 0.9179          & 0.9220  & 0.9251 & 0.9220 & 0.9326 & 0.9317   \\
CBL~\cite{Cui2019Class} ($\beta\!=\!0.9999$)              & 0.7885          & 0.8099          & 0.8977          & 0.9127          & 0.9241          & 0.9226   & 0.9261 & 0.9320 & 0.9307 & 0.9376  \\
\hline
(ours) ALT Mode              & \textbf{0.8240} & 0.8520          & 0.8900          & 0.8880          & 0.8520          & 0.8920 &0.8860 &0.8880 &0.8840 &0.8760     \\
(ours) AUG Mode               & 0.8060          & \textbf{0.8740} & \textbf{0.9140} & 0.9160 & \textbf{0.9340} & 0.9220   &0.9140  &0.9280 &0.9340 &0.9340 \\
\hline
\end{tabular}}
\vspace{2mm}
\caption {Comparison of the proposed method on the CelebA dataset using both AUG and ALT mode. The number of majority class training images (female) in CelebA is fixed to 900. Here we provide the \textbf{$F_1$ score of the minority class (Male)} for all the imbalance ratios starting from 900:50 (Female:Male) to a perfectly balanced scenario of 900:900 (Female:Male) images in the training set. The $F_1$ score for imabalance ratio till 900:500 (Female:Male) is replicated from the main paper for completeness.
Our proposed approach outperforms all others comprehensively in the high imbalance regions. The additional ratios show that the traditional methods start to do well as the dataset is more and more balanced. However, our augmented way of training is still very good and the difference in performance is only at the third decimal place in the high 90's.
}
\label{tab:celeba_male_f1}
\end{table*}
\begin{table*}[htbp]
\centering
\resizebox{\linewidth}{!}{ 
\begin{tabular}{l||cccccccccc}
\hline
\textbf{Dataset} & \multicolumn{8}{c}{\textbf{CelebA}}\\
\hline
\#Male                       & 50      & 100     & 200     & 300     & 400     & 500     & 600   &700 &800 &900       \\
\hline
\hline
Vanilla                      & 0.6840  & 0.7560  & 0.8380  & 0.8740  & 0.9120  & 0.9240  & 0.9300  &0.9280 &0.9340 &\textbf{0.9380}       \\
\hline
US                & 0.8261 & 0.8610 & 0.9043 &\textbf{ 0.9170} & 0.9192 & 0.9293 & 0.9217 & 0.9141 & 0.9168 & 0.9210  \\
TS       & 0.6860  & 0.7840  & 0.8440  & 0.8740  & 0.9080  & 0.9220  & 0.9300   & 0.9320  & 0.9340  &\textbf{0.9380}     \\
CS               & 0.8301  & 0.8436  & 0.9060  & 0.9148  & 0.9256  & 0.9255  & 0.9243 & 0.9286  & 0.9250  & 0.9334   \\
OS        & 0.7691  & 0.8170  & 0.8885  & 0.9080  & 0.9218  & 0.9198  & 0.9238      & 0.9301  & 0.9319  & 0.9331        \\
US + CS & 0.7936  & 0.8643  & 0.8986  & 0.9122  & 0.9147  & 0.9172  & 0.9234     & 0.9249  & 0.9277  & 0.9209        \\
OS + CS & 0.8268  & 0.8513  & 0.8973  & 0.9119  & 0.9230  & \textbf{0.9259}  & \textbf{0.9323}     & 0.9320  & 0.9303  & 0.9380          \\
SMOTE  ~\cite{Chawla2002Smote}                       & 0.7774  & 0.8288  & 0.8936  & 0.8981  & 0.9192  & 0.9218  & 0.9294  & 0.9363  &\textbf{0.9379}  & 0.9356     \\
CBL~\cite{Cui2019Class} ($\beta\!=\!0.9$)                 & 0.7963  & 0.8372  & 0.8962  & 0.9104  & 0.9147  & 0.9220  & 0.9260 &\textbf{ 0.9370}  & 0.9284  & 0.9376       \\
CBL~\cite{Cui2019Class} ($\beta\!=\!0.99$)                & 0.8011  & 0.8483  & 0.9017  & 0.9161  & 0.9195  & 0.9247  & 0.9229  & 0.9284  & 0.9338  & 0.9366      \\
CBL~\cite{Cui2019Class} ($\beta\!=\!0.999$)               & 0.8218  & 0.8587  & 0.9031  & 0.9164  & 0.9186  & 0.9252  & 0.9256  &0.9246  & 0.9328  & 0.9296      \\
CBL~\cite{Cui2019Class} ($\beta\!=\!0.9999$)              & 0.8322  & 0.8494  & 0.9047  & 0.9111  & 0.9238  & 0.9240  & 0.9278  & 0.9306  & 0.9306  & 0.9370   \\
\hline
ALT Mode    &0.8300             & 0.8580  & 0.8840  & 0.8820  & 0.8380  & 0.8940  & 0.8820  &0.8820 &0.8860 &0.8780    \\
AUG Mode                 &\textbf{ 0.8360}  & \textbf{0.8820}  & \textbf{0.9160}  & 0.9160  &\textbf{ 0.9300}  & 0.9200  & 0.9160 &0.9260 &0.9320 &0.9320    \\ 
\hline
\end{tabular}}
\vspace{2mm}
\caption{Comparison of the proposed method on the CelebA dataset using both AUG and ALT mode. The number of majority class training images (female) in CelebA is fixed to 900. Here we provide the \textbf{$F_1$ score of the majority class (Female)} for all the imbalance ratios starting from 900:50 (Female:Male) to a perfectly balanced scenario of 900:900 (Female:Male) images in the training set. We can see that our proposed approach is comprehensively outperforming the other methods in the highly imbalanced region (left part of the table) and performing at par with the other methods in the more balanced scenario (towards right of the table). While Table~\ref{tab:celeba_male_f1} shows that the minority class (male) classification gets improved due to our Cycle-GAN based dataset repairment, it also improves the classification performance of the majority class (female).
}
\label{tab:celebA_female_f1}
\end{table*}
\begin{table*}[htbp]
\centering
\resizebox{\linewidth}{!}{ 
\begin{tabular}{l||cccccccccc}
\hline
\textbf{Dataset} & \multicolumn{8}{c}{\textbf{CelebA}}\\
\hline
\#Male                       & 50              & 100             & 200             & 300             & 400             & 500             & 600  &700 &800 &900  \\
\hline
\hline
Vanilla &0.5400 &0.6760 &0.8120 &0.86400 &0.9090 &0.9210 &0.9290 &0.9280 &0.9350 &0.9360\\
\hline
US                & 0.8153          & 0.8553          & 0.9040          & \textbf{0.9173} & 0.9187          & \textbf{0.9300} & 0.9220    & 0.9147 & 0.9193 & 0.9223       \\
TS                 & 0.5430          & 0.7290          & 0.8210          & 0.8590          & 0.9030          & 0.9200          & \textbf{0.9300}   & 0.9300 & 0.9350 & 0.9360 \\
CS              & 0.8097          & 0.8250          & 0.9020          & 0.9143          & 0.9253          & 0.9250          & 0.9247         & 0.9277 & 0.9253 & 0.9347  \\
OS                 & 0.7030          & 0.7830          & 0.8823          & 0.9060          & 0.9200          & 0.9193          & 0.9227       & 0.9293 & 0.9320 & 0.9330    \\
US + CS & 0.8003          & 0.8560          & 0.9003          & 0.9143          & 0.9170          & 0.9197          & 0.9247         & 0.9263 & 0.9300 & 0.9230     \\
OS + CS  & 0.8010          & 0.8393          & 0.8947          & 0.9093          & 0.9227          & 0.9260          & 0.9317          & 0.9317 & 0.9300 & \textbf{0.9387} \\
SMOTE  ~\cite{Chawla2002Smote}                       & 0.7197          & 0.8033          & 0.8877          & 0.8940          & 0.9180          & 0.9213          & 0.9290   & 0.9357 &\textbf{0.9383} & 0.9357       \\
CBL~\cite{Cui2019Class} ($\beta\!=\!0.9$)                & 0.7493          & 0.8123          & 0.8917          & 0.9083          & 0.9133          & 0.9213          & 0.9260   &\textbf{ 0.9373} & 0.9283 & 0.9380       \\
CBL~\cite{Cui2019Class} ($\beta\!=\!0.99$)                & 0.7613          & 0.8287          & 0.8980          & 0.9140          & 0.9187          & 0.9237          & 0.9210   & 0.9283 & 0.9343 & 0.9373        \\
CBL~\cite{Cui2019Class} ($\beta\!=\!0.999$)               & 0.7990          & 0.8437          & 0.8980          & 0.9143          & 0.9183          & 0.9237          & 0.9253    & 0.9233 & 0.9327 & 0.9307     \\
CBL~\cite{Cui2019Class} ($\beta\!=\!0.9999$)              & 0.8130          & 0.8320          & 0.9013          & 0.9120          & 0.9240          & 0.9233          & 0.9270   & 0.9313 & 0.9307 & 0.9373       \\
\hline
ALT Mode                 & \textbf{0.8260} & 0.8540          & 0.8880          & 0.8870          & 0.8460          & 0.8930          & 0.8820 &0.8860 &0.8850 &0.8790          \\
AUG Mode                  & 0.8230          & \textbf{0.8800} & \textbf{0.9180} & 0.9160          & \textbf{0.9300} & 0.9210          & 0.9150   &0.9260 &0.9350 &0.9360      \\
\hline
\end{tabular}}
\vspace{2mm}
\caption{Comparison of the proposed method on the CelebA dataset using both AUG and ALT mode in terms of \textbf{Average Class Specific Accuracy (ASCA)}. The class specific accuracy is averaged over the two classes (Male and Female). The number of majority class training images (Female) in CelebA is fixed to 900. We provide the ACSA for all the imbalance ratios starting from 900:50 (Female:Male) to a perfectly balanced scenario of 900:900 (Female:Male) images in the training set. We can see that our proposed approach is comprehensively outperforming the other methods in the highly imbalanced region (left part of the table) and performing at par with the other methods in the more balanced scenario (towards right of the table). The ACSA values show that the recall values for both the classes increase on average as a result of the application of imbalance mitigation strategies but our proposed method works best especially in presence of highly skewed training data.
}
\label{tab:celeba_acsa}
\end{table*}
\begin{table}[htbp]
\setlength\tabcolsep{14pt}
\centering
\resizebox{\linewidth}{!}{ 
\begin{tabular}{l||llll}
\hline
\textbf{Dataset} & \multicolumn{3}{c}{\textbf{Horse2Zebra}}\\
\hline
\#Zebra                      & 25      & 50     & 75     & 100    \\
\hline
\hline
Vanilla                      & 0.6740       & 0.7680       & 0.8200       & 0.8400       \\
\hline
US                & 0.5333       & 0.8217       & 0.8784       & 0.8741       \\
TS                 & 0.6720       & 0.7860       & 0.8160       & 0.8540       \\
CS               & 0.8469       & 0.8692       & 0.8919       & 0.8873       \\
OS                 & 0.7452       & 0.8559       & 0.8742       & 0.8821       \\
US + CS & 0.4410 & 0.8097 & 0.8880 & \textbf{0.8912} \\
OS + CS  & 0.8282       & 0.8779       & 0.8910       & 0.8892       \\
SMOTE ~\cite{Chawla2002Smote}                        & 0.6964       & 0.8205       & 0.8511       & 0.8684       \\
CBL~\cite{Cui2019Class} ($\beta\!=\!0.9$)                 & 0.8373       & 0.8658       & 0.8856       & 0.8790       \\
CBL~\cite{Cui2019Class} ($\beta\!=\!0.99$)                & 0.8537       & 0.8643       & 0.8816       & 0.8856       \\
CBL~\cite{Cui2019Class} ($\beta\!=\!0.999$)               & 0.8569       & 0.8585       & 0.8793       & 0.8862       \\
CBL~\cite{Cui2019Class} ($\beta\!=\!0.999$)             & 0.8560       & 0.8668       & 0.8862       & 0.8868       \\
\hline
ALT Mode                & 0.8380       & 0.8120       & 0.7880       & 0.8120       \\
AUG Mode                  & \textbf{0.8800 }     &\textbf{ 0.8900}      & \textbf{0.8980 }     & 0.8760      \\
\hline
\end{tabular}}
\vspace{2mm}
\caption{Comparison of the proposed method on the Horse2Zebra dataset using both AUG and ALT mode. The number of majority class training images (Horse) is fixed to 450. Here we provide the \textbf{$F_1$ score of the majority class (Horse)} for all the imbalance ratios starting from 450:25 (Horse:Zebra) to a perfectly balanced scenario of 450:100 (Horse:Zebra) images in the training set. We can see that our proposed approach is comprehensively outperforming the other methods in the highly imbalanced region (left part of the table) and performing at par with the other methods in the more balanced scenario (towards right of the table). While Table 3 in the main paper shows that the minority class (zebra) classification gets improved due to our Cycle-GAN based dataset repairment, it also improves the classification performance of the majority class (horse).
}
\label{tab:horses_f1}
\end{table}
\begin{table}[htbp]
\setlength\tabcolsep{14pt}
\centering
\resizebox{\linewidth}{!}{ 
\begin{tabular}{l||cccc}
\hline
\textbf{Dataset} & \multicolumn{3}{c}{\textbf{Horse2Zebra}}\\
\hline
\#Zebra                       & 25             & 50             & 75            & 100                   \\
\hline
\hline
Vanilla &0.5150 &0.6940 &0.7800 &0.8100 \\
\hline
US                & 0.5000            & 0.8080          & 0.8640         & 0.8640                 \\
TS                 & 0.5090          & 0.7290          & 0.7740         & 0.8290                 \\
CS               & 0.8240          & 0.8533   & 0.8820         & \textbf{0.8787} \\
OS                & 0.65733   & 0.8333   & 0.8580         & 0.87330         \\
US + CS & 0.5133   & 0.7800           & 0.8773 & 0.8820                 \\
OS + CS  & 0.7927   & 0.8667   & 0.8807  & 0.8820                 \\
SMOTE     ~\cite{Chawla2002Smote}                    & 0.5627   & 0.7847   & 0.8260         & 0.8500                  \\
CBL~\cite{Cui2019Class} ($\beta\!=\!0.9$)                 & 0.8067   & 0.8520          & 0.8727  & 0.8640                 \\
CBL~\cite{Cui2019Class} ($\beta\!=\!0.99$)               & 0.8353   & 0.85           & 0.8713 & 0.8740                 \\
CBL~\cite{Cui2019Class} ($\beta\!=\!0.999$)               & 0.8433   & 0.8427   & 0.8673  & 0.8780                 \\
CBL~\cite{Cui2019Class} ($\beta\!=\!0.9999$)              & 0.8380          & 0.8467   & 0.8760         & 0.8773          \\
\hline
ALT Mode                 & 0.8240              & 0.8230          & 0.7980         & 0.8160                 \\
AUG Mode                  & \textbf{0.8670} & \textbf{0.8770} & \textbf{0.8900} & 0.8630                 \\
\hline
\end{tabular}}
\vspace{2mm}
\caption{Comparison of the proposed method on the Horse2Zebra dataset using both AUG and ALT mode in terms of \textbf{Average Class Specific Accuracy (ASCA)}. The class specific accuracy is averaged over the two classes (Horse and Zebra). The number of majority class training images (Horse) is fixed to 450. We provide the ACSA for all the imbalance ratios starting from 450:25 (Horse:Zebra) to 450:100 (Horse:Zebra) images in the training set. We can see that our proposed approach is comprehensively outperforming the other methods. The ACSA values show that the recall values for both the classes increase on average as a result of the application of imbalance mitigation strategies but our proposed method works best especially in presence of highly skewed training data.
}
\label{tab:horse2zebra_acsa}
\end{table}
\begin{table*}[htbp]
\setlength\tabcolsep{12pt}
\centering
\resizebox{1\textwidth}{!}{ 
\begin{tabular}{l||ccccccc}
\hline
\textbf{Dataset} & \multicolumn{6}{c}{\textbf{CUB-200-2011}}\\
\hline
\#Flycatcher                 & 12              & 25                  & 50              & 75              & 100             & 125               \\
\hline
\hline
Vanilla                      & 0.6700          & 0.6800              & 0.7020          & 0.7020          & 0.7340          & 0.7380  \\
\hline
US      & 0.4000         & 0.6694            & 0.6474         & 0.6803         & 0.6470         & 0.6113        \\
TS                 & 0.6720          & 0.6800              & 0.7020          & 0.7060          & 0.6915          & 0.7320    \\
CS      & \textbf{0.6800} & 0.7134              & 0.6994          & 0.7027          & 0.7373          & 0.7184            \\
OS                 & 0.6721          & 0.6916              & 0.7100          & 0.7174          & 0.7261          & 0.7221        \\
US + CS & 0.3058          & 0.6033              & 0.6287    & 0.6573          & 0.6116          & 0.6650       \\
OS + CS  & 0.5712          & 0.6784              & 0.6965          & 0.6808          & 0.7108          & 0.7136       \\
SMOTE  ~\cite{Chawla2002Smote}                       & 0.6685          & 0.6949              & 0.6123          & 0.6478          & 0.7334          & 0.7250        \\
CBL~\cite{Cui2019Class} ($\beta\!=\!0.9$)                & 0.6695          & 0.6028 & 0.7217          & \textbf{0.7317} & 0.7228          & \textbf{0.7381}         \\
CBL~\cite{Cui2019Class} ($\beta\!=\!0.99$)                & 0.6763          & 0.7080              & \textbf{0.7424} & 0.7218          & \textbf{0.7439} & 0.7102           \\
CBL~\cite{Cui2019Class} ($\beta\!=\!0.999$)               & 0.6591          & \textbf{0.7164}     & 0.7149          & 0.7073          & 0.7036          & 0.7210            \\
CBL~\cite{Cui2019Class} ($\beta\!=\!0.999$)              & 0.6523          & 0.7094              & 0.7058          & 0.7142          & 0.7387          & 0.7179              \\
\hline
ALT Mode                 & 0.6260          & 0.6180              & 0.6680          & 0.6160          & 0.6640          & 0.6200      \\
AUG Mode                  & 0.6180         & 0.6660             & 0.6560         & 0.7100         & 0.6900        & 0.7180    \\
\hline
\end{tabular}}
\vspace{2mm}
\caption{Comparison of the proposed method on the CUB-200-2011 dataset using both AUG and ALT mode. The number of majority class training images (sparrow) is fixed to 250. Here we provide the \textbf{$F_1$ score of the majority class (Sparrow)} for all the imbalance ratios starting from 250:12 (Sparrow:Flycatcher) to 250:125 (Sparrow:Flycatcher) images in the training set. We can see that our proposed approach is not doing good in recovering the $F_1$ score of the majority class in this dataset. This may be due to the presence of very few images for both the classes and also due to the difficult finegrained nature of the task in this particular dataset. The effect can also be seen in Table~\ref{tab:acsa_cub} where the ACSA is just at par for these ratios for the proposed method. However in Table 1 of the main paper we see that the F1 score for the other class (Flycather) is always high for our proposed method.
}
\label{tab:cub_sparrow_f1}
\end{table*}
\begin{table*}[htbp]
\setlength\tabcolsep{12pt}
\centering
\resizebox{\textwidth}{!}{ 
\begin{tabular}{l||ccccccccccccc}
\hline
\textbf{Dataset} & \multicolumn{6}{c}{\textbf{CUB-200-2011}}\\
\hline
\#Flycatcher                 & 12     & 25     & 50     & 75     & 100    & 125    \\
\hline
\hline
Vanilla                      & 0.5060 & 0.5320 & 0.5800 & 0.5960 & 0.6660 & 0.6900 \\
\hline
US               & 0.5040 & 0.5980 & 0.6340 & 0.6860 & 0.6880 & 0.6680 \\
TS                 & 0.5060 & 0.5320 & 0.5800 & 0.5980 & 0.6101 & 0.6840 \\
CS              & 0.5800 & 0.6400 & 0.6300 & 0.6800 & 0.7000 & 0.7200 \\
OS                & 0.5140 & 0.5540 & 0.6080 & 0.6420 & 0.6860 & 0.6840 \\
US + CS & 0.5380 & 0.5780 & 0.6520 & 0.6620 & 0.6640 & 0.6960 \\
OS + CS  & 0.5440 & 0.5960 & 0.6500 & 0.6520 & 0.6740 & 0.7040 \\
SMOTE     ~\cite{Chawla2002Smote}                    & 0.5100 & 0.6200 & 0.6460 & 0.6280 & 0.7060 & 0.7020 \\
CBL~\cite{Cui2019Class} ($\beta\!=\!0.9$)                   & 0.5220 & 0.6056 & 0.6460 & 0.6700 & 0.6800 & 0.7180 \\
CBL~\cite{Cui2019Class} ($\beta\!=\!0.99$)                 & 0.5640 & 0.6440 &\textbf{0.6880} & 0.6800 & 0.7000 & 0.6900 \\
CBL~\cite{Cui2019Class} ($\beta\!=\!0.999$)                 & 0.5600 & 0.6460 &0.6760 & 0.6440 & 0.6580 & 0.7220 \\
CBL~\cite{Cui2019Class} ($\beta\!=\!0.9999$)                & 0.5520 & \textbf{0.6640} & 0.6360 & 0.6660 & \textbf{0.7040} & \textbf{0.7280} \\
\hline
ALT Mode                 & 0.5800 & 0.5800 & 0.6280 & 0.6320 & 0.6800 & 0.6160 \\
AUG Mode                  & \textbf{0.6080} & 0.6400 & 0.6620 & \textbf{0.7080} & 0.6980 & 0.7180 \\
\hline                             
\end{tabular}}
\vspace{2mm}
\caption{Comparison of the proposed method on the CUB-200-2011 dataset using both AUG and ALT mode in terms of \textbf{Average Class Specific Accuracy (ASCA)}. The class specific accuracy is averaged over the two classes (Sparrow and Flycatcher). The number of majority class training images (Sparrow) is fixed to 250. We provide the ACSA for all the imbalance ratios starting from 250:12 (Sparrow:Flycatcher) to 250:125 (Sparrow:Flycatcher) images in the training set. We can see that our proposed approach is better or at par with the other methods.
}
\label{tab:acsa_cub}
\end{table*}

\begin{table*}[htbp]
\setlength\tabcolsep{2pt}
\centering
\begin{tabular}{|c||c|c||c|c|}
\hline
  \textbf{Real}&
 \includegraphics[width=1in]{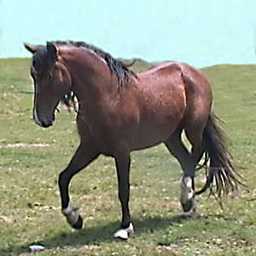} &
 \includegraphics[width=1in]{figs/figsH2Z/zebra/real_horse.jpg} &
 \includegraphics[width=1in]{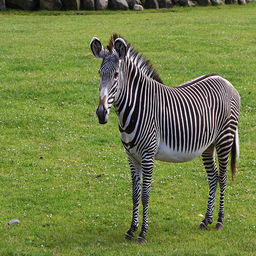} &
 \includegraphics[width=1in]{figs/figsH2Z/horse/real.jpg} 
 \\
 \hline
 \hline

  \textbf{$\gamma$}&
 \textbf{CycleGAN} &
 \textbf{ALT-Mode } &
 \textbf{CycleGAN } &
 \textbf{ALT-Mode }\\
  ~ &\textbf{Zebra} &
\textbf{Zebra }&
\textbf{Horse} &
\textbf{Horse}\\
 \hline
 \textbf{450/25}&
 \includegraphics[width=1in]{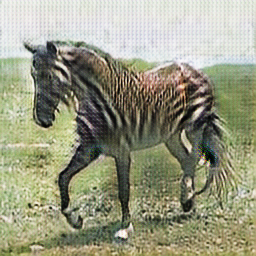} &
 \includegraphics[width=1in]{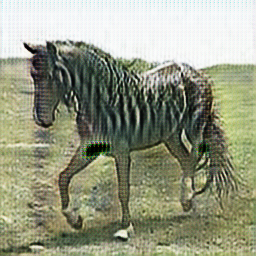} &
 \includegraphics[width=1in]{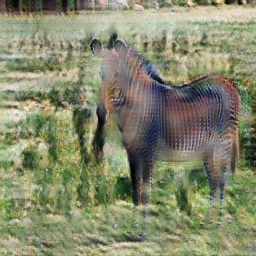} &
 \includegraphics[width=1in]{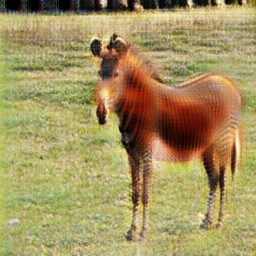} \\
 \hline
 \textbf{450/50}&
 \includegraphics[width=1in]{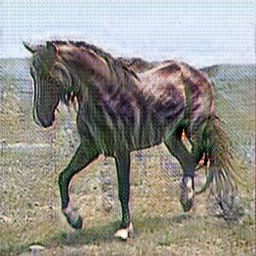} &
 \includegraphics[width=1in]{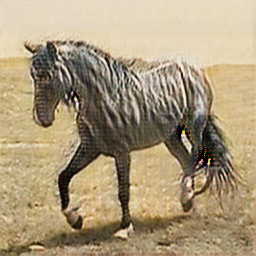} &
 \includegraphics[width=1in]{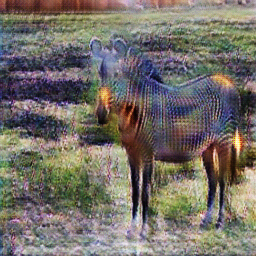} &
 \includegraphics[width=1in]{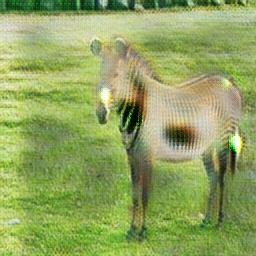} \\
 \hline
 \textbf{450/75}&
 \includegraphics[width=1in]{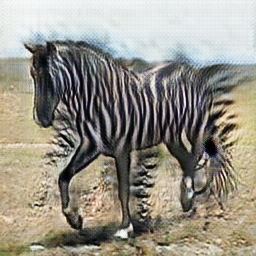} &
 \includegraphics[width=1in]{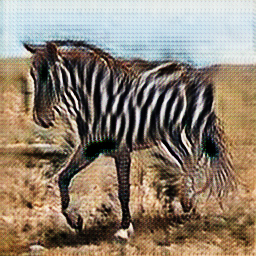} &
 \includegraphics[width=1in]{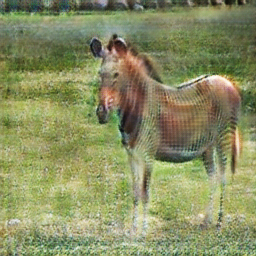} &
 \includegraphics[width=1in]{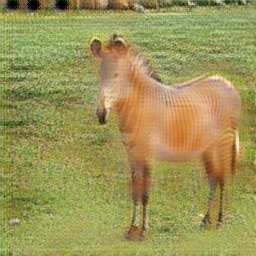} \\
 \hline
  \textbf{450/100}&
 \includegraphics[width=1in]{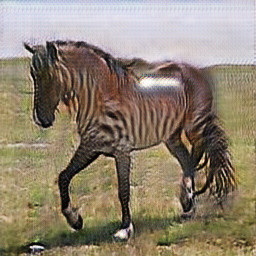} &
 \includegraphics[width=1in]{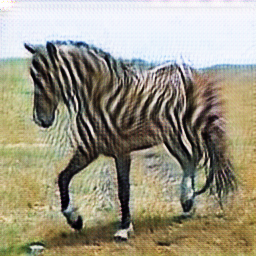} &
 \includegraphics[width=1in]{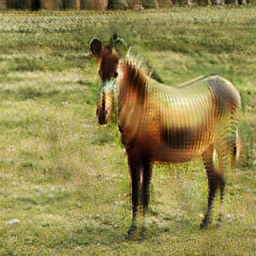} &
 \includegraphics[width=1in]{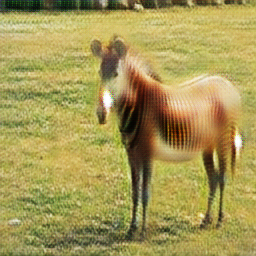} \\
 \hline
\end{tabular}
\vspace{2mm}
\caption{Comparison of generated images from only CycleGAN and our ALT Mode trained model for Horse2Zebra dataset. $\gamma$ indicates the imabalance ratio which is defined as the number of training examples in the majority class to the same in the minority class. Here horse is the majority class and zebra is the minority class. The top row indicates the input images to the generators of both cycleGAN and ALT mode trained model. Subsequently, every row indicates the generated images from both the models respectively for the imbalance ratio mentioned in the leftmost column.}
\label{tab:horse2zebrafig}
\end{table*}

\begin{table*}[htbp]
\setlength\tabcolsep{2pt}
\centering
\begin{tabular}{|c||c|c||c|c|}
\hline
  \textbf{Real}&
\includegraphics[width=1in,height=0.85in]{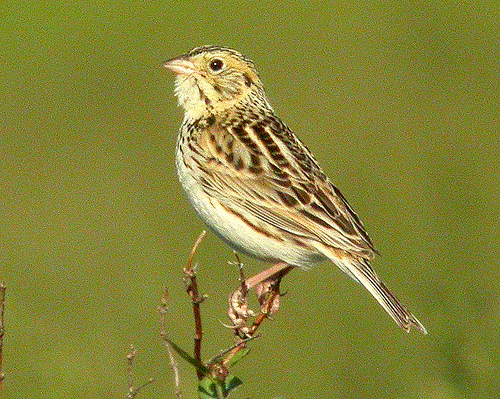} &
\includegraphics[width=1in,height=0.85in]{figs/figsCUB/flycatcher/Baird_Sparrow_0023_3046631067.jpg} &
\includegraphics[width=1in,height=0.85in]{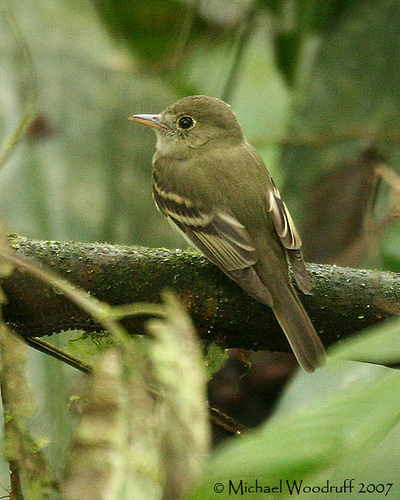} &
\includegraphics[width=1in,height=0.85in]{figs/figsCUB/sparrow/Acadian_Flycatcher_0012_443003714.jpg}
\\
\hline
\hline
 \textbf{$\gamma$}&
\textbf{CycleGAN } &
\textbf{ALT-Mode }&
\textbf{CycleGAN } &
\textbf{ALT-Mode }\\
~ &\textbf{Flycatcher} &
\textbf{Flycatcher }&
\textbf{Sparrow} &
\textbf{Sparrow}\\
\hline
 \textbf{250/12}&
\includegraphics[width=1in,height=0.85in]{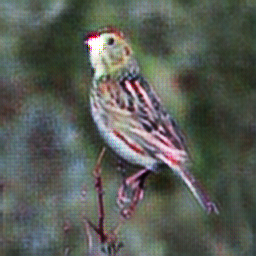} &
\includegraphics[width=1in,height=0.85in]{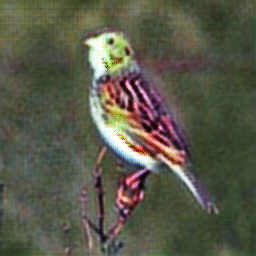} &
\includegraphics[width=1in,height=0.85in]{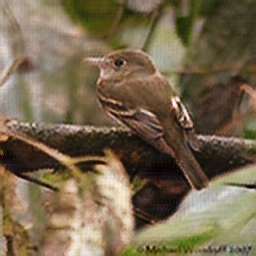} &
\includegraphics[width=1in,height=0.85in]{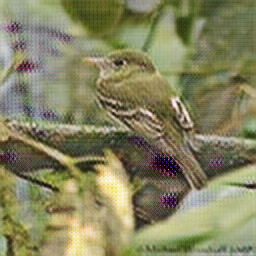} \\
\hline
 \textbf{250/25}&
\includegraphics[width=1in,height=0.85in]{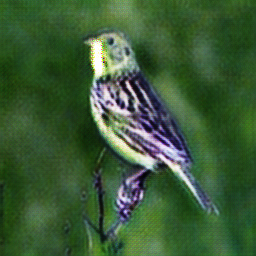} &
\includegraphics[width=1in,height=0.85in]{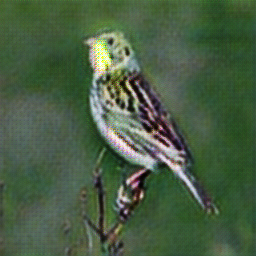} &
\includegraphics[width=1in,height=0.85in]{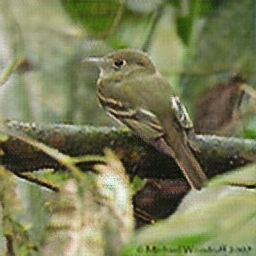} &
\includegraphics[width=1in,height=0.85in]{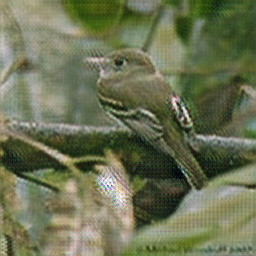} \\
\hline
 \textbf{250/50}&
\includegraphics[width=1in,height=0.85in]{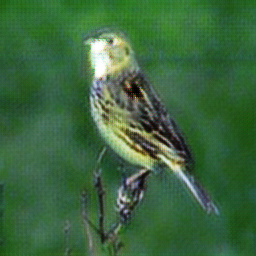} &
\includegraphics[width=1in,height=0.85in]{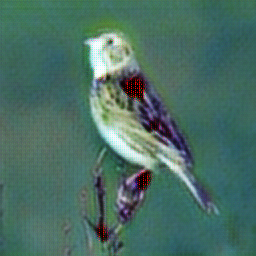} &
 \includegraphics[width=1in,height=0.85in]{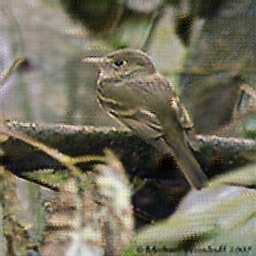} &
\includegraphics[width=1in,height=0.85in]{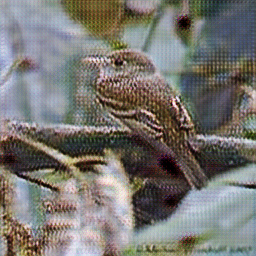} \\
\hline
  \textbf{250/75}&
\includegraphics[width=1in,height=0.85in]{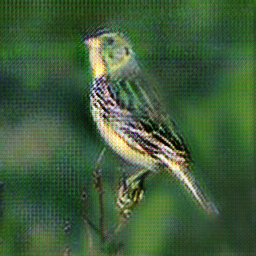} &
\includegraphics[width=1in,height=0.85in]{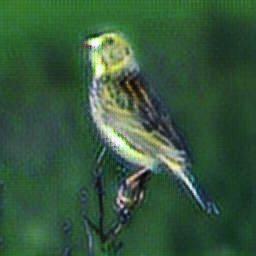} &
 \includegraphics[width=1in,height=0.85in]{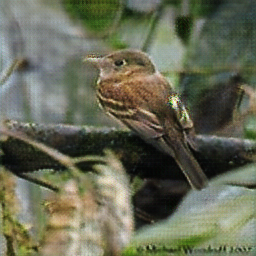} &
\includegraphics[width=1in,height=0.85in]{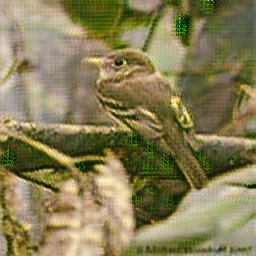} \\
\hline
 \textbf{250/100}&
\includegraphics[width=1in,height=0.85in]{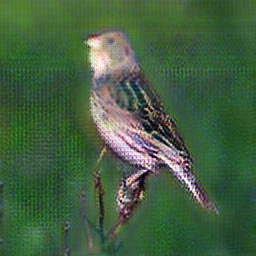} &
\includegraphics[width=1in,height=0.85in]{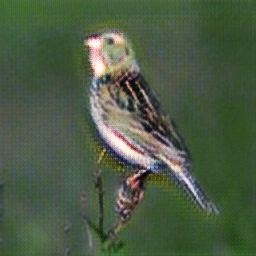} &
\includegraphics[width=1in,height=0.85in]{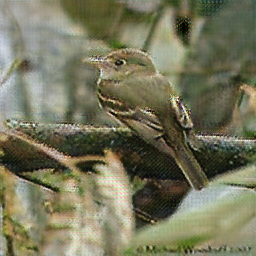} &
\includegraphics[width=1in,height=0.85in]{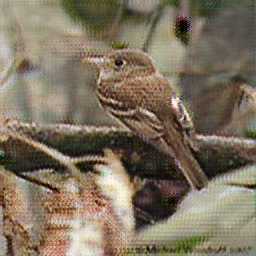} \\
\hline
\textbf{250/125}&
\includegraphics[width=1in,height=0.85in]{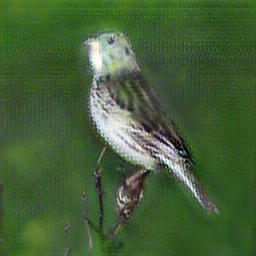} &
\includegraphics[width=1in,height=0.85in]{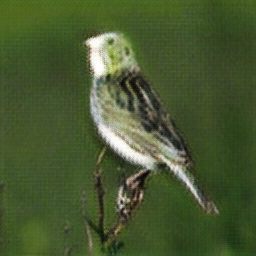} &
\includegraphics[width=1in,height=0.85in]{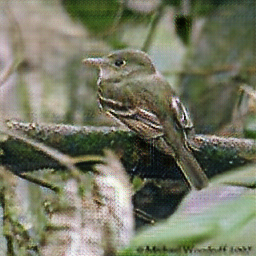} &
\includegraphics[width=1in,height=0.85in]{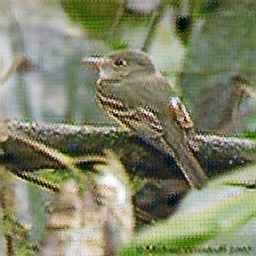} \\
\hline
\end{tabular}
\vspace{2mm}
\caption{Comparison of generated images from only CycleGAN and our ALT Mode trained model for Fine grainedCUB dataset. $\gamma$ indicates the imabalance ratio which is defined as the number of training examples in the majority class to the same in the minority class. Here sparrow is the majority class while flycatcher is the minority class. The top row indicates the input images to the generators of both cycleGAN and ALT mode trained model. Subsequently, every row indicates the generated images from both the models respectively for the imbalance ratio mentioned in leftmost column.}
\label{tab:CUBfig}
\end{table*}

\begin{table*}[htbp]
\setlength\tabcolsep{2pt}
\centering
\begin{tabular}{|c|c|c|c|}
\multicolumn{4}{c}{\textbf{RISE} heatmaps for \textbf{Horse2Zebra} dataset}\\
\hline

 Original Image &
 Vanilla Mode &
 ALT Mode &
 AUG Mode\\
 \hline

\includegraphics[width=1in, height=1in]{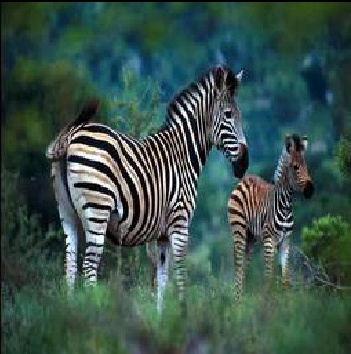}&
\includegraphics[width=1in]{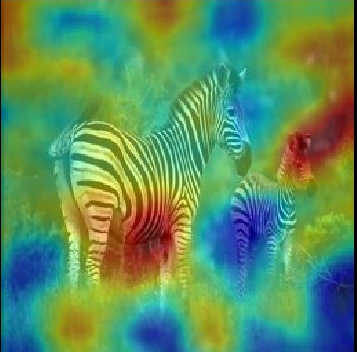}&
\includegraphics[width=1in]{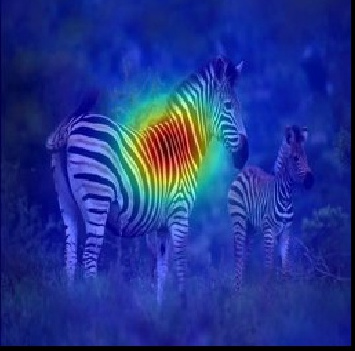} &
\includegraphics[width=1in]{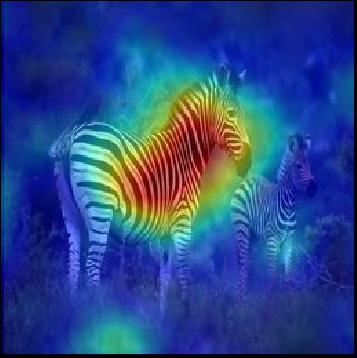} \\

\includegraphics[width=1in, height=1in]{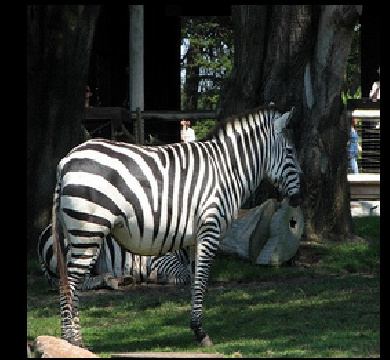}&
\includegraphics[width=1in]{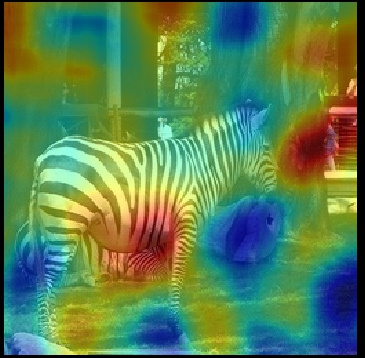}&
\includegraphics[width=1in]{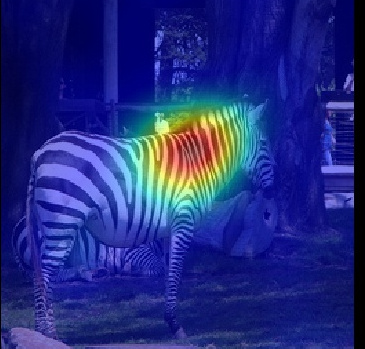} &
\includegraphics[width=1in]{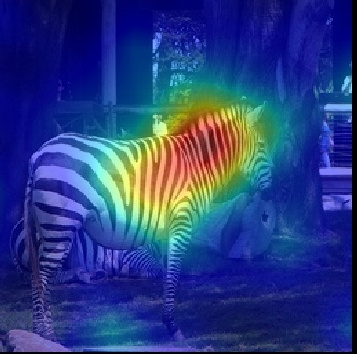} \\

\includegraphics[width=1in, height=1in]{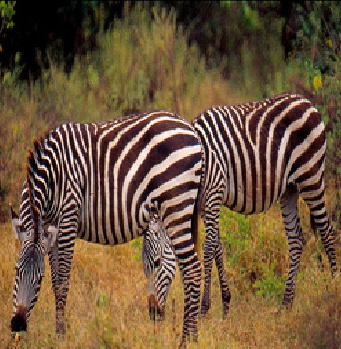}&
\includegraphics[width=1in]{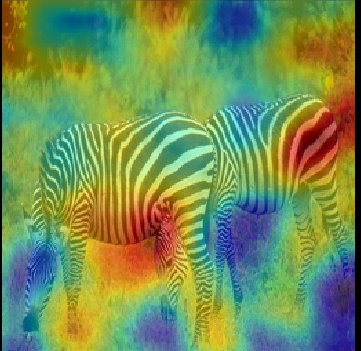}&
\includegraphics[width=1in]{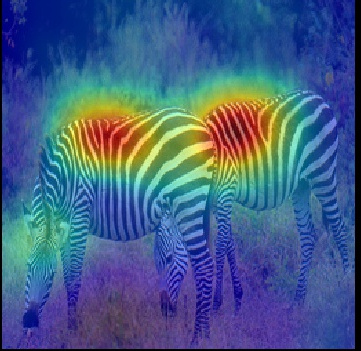} &
\includegraphics[width=1in]{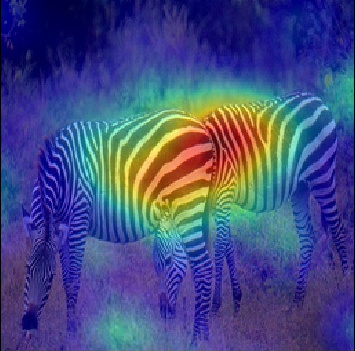} \\

\includegraphics[width=1in, height=1in]{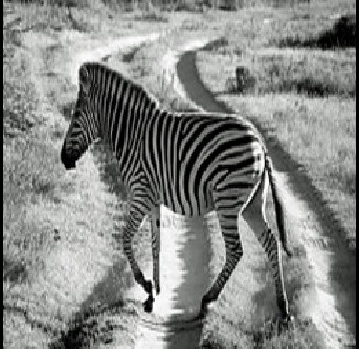}&
\includegraphics[width=1in]{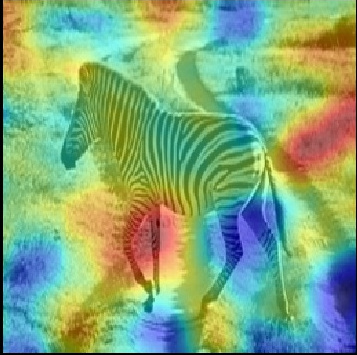}&
\includegraphics[width=1in]{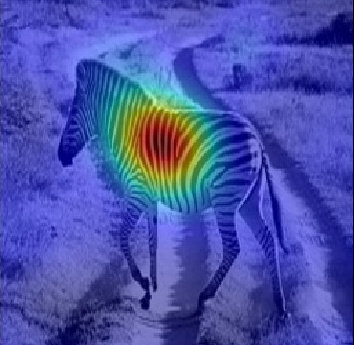} &
\includegraphics[width=1in]{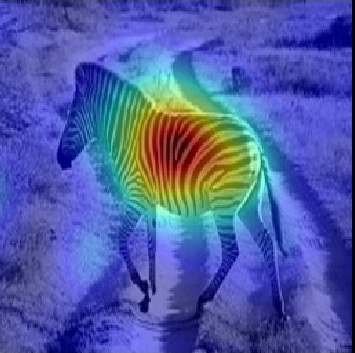} \\

\includegraphics[width=1in, height=1in]{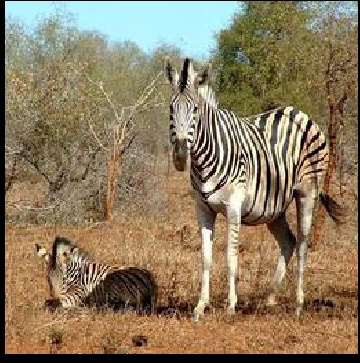}&
\includegraphics[width=1in]{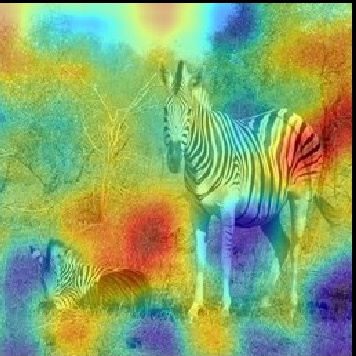}&
\includegraphics[width=1in]{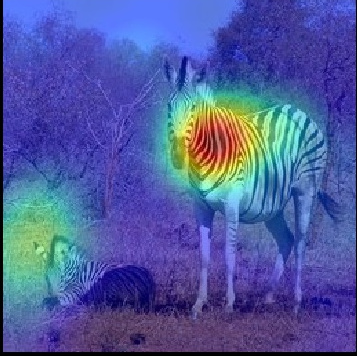} &
\includegraphics[width=1in]{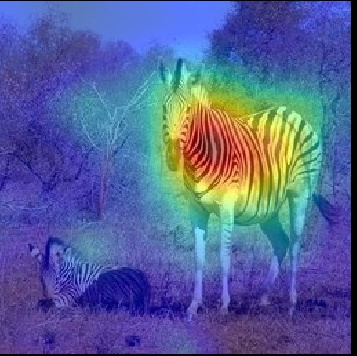} \\

\hline

\end{tabular}

\vspace{2mm}
\caption{\textbf{RISE} heatmaps of images from \textbf{Horse2Zebra} dataset. In each of the rows the leftmost  column (col 1)  shows  the  original image of the Zebra while  col 2-4 show  RISE heatmaps for Vanilla, ALT and AUG Modes respectively (importance increases from blue to red). The column sub-headings denote the corresponding classifier. Note that the heatmaps are generated for the predicted output by the classifiers. Improvement in the explanations can be observed for the ALT and AUG Mode classifiers as compared to the corresponding Vanilla classifier.}
\label{tab:rise_h2z}

\end{table*}

\begin{table*}[htbp]
\setlength\tabcolsep{2pt}
\centering
\begin{tabular}{|c|c|c|c|}
\multicolumn{4}{c}{\textbf{RISE} heatmaps for \textbf{CUB} dataset}\\
\hline

 Original Image &
 Vanilla Mode &
 ALT Mode &
 AUG Mode\\
 \hline

\includegraphics[width=1in, height=1in]{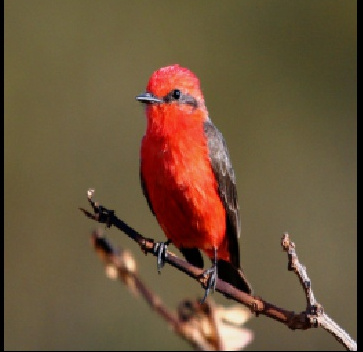}&
\includegraphics[width=1in]{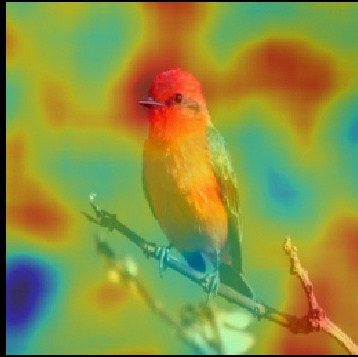}&
\includegraphics[width=1in]{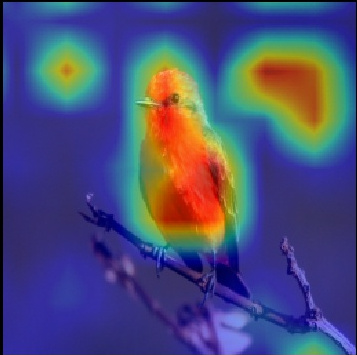} &
\includegraphics[width=1in]{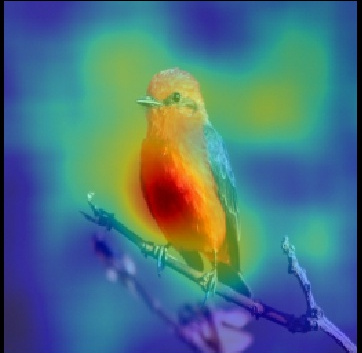} \\

\includegraphics[width=1in, height=1in]{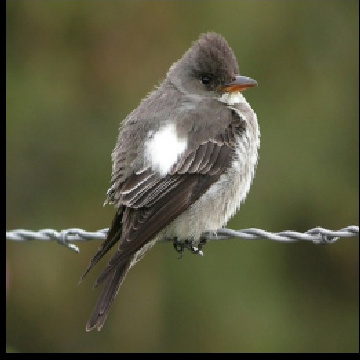}&
\includegraphics[width=1in]{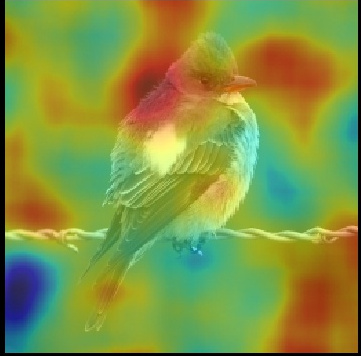}&
\includegraphics[width=1in]{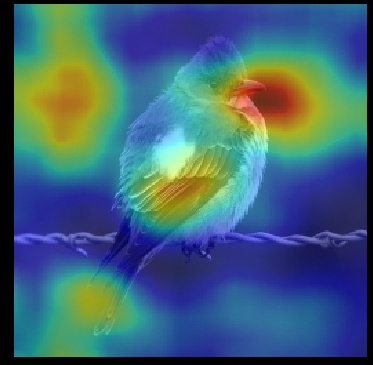} &
\includegraphics[width=1in]{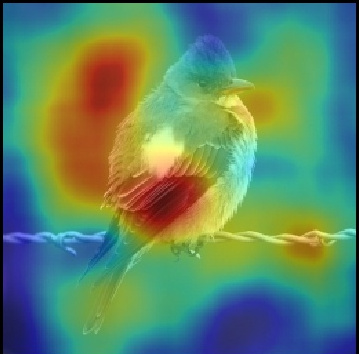} \\

\includegraphics[width=1in, height=1in]{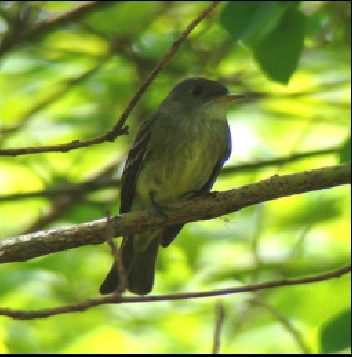}&
\includegraphics[width=1in]{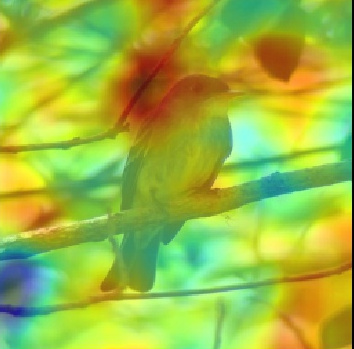}&
\includegraphics[width=1in]{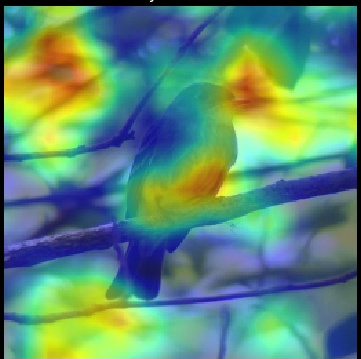} &
\includegraphics[width=1in]{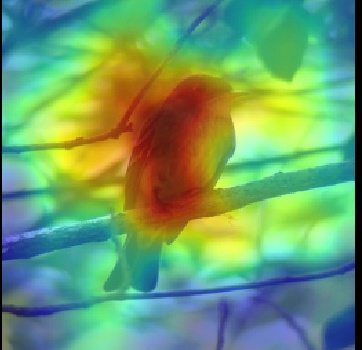} \\

\includegraphics[width=1in, height=1in]{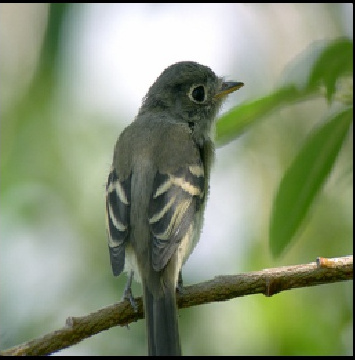}&
\includegraphics[width=1in]{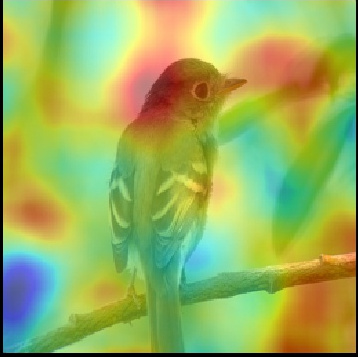}&
\includegraphics[width=1in]{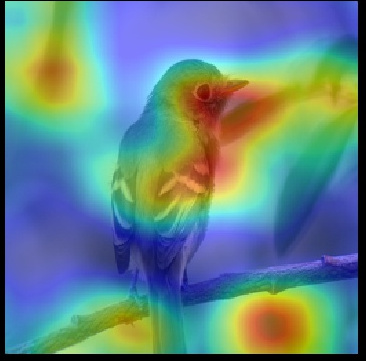} &
\includegraphics[width=1in]{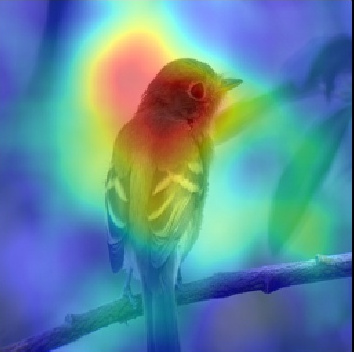} \\

\includegraphics[width=1in, height=1in]{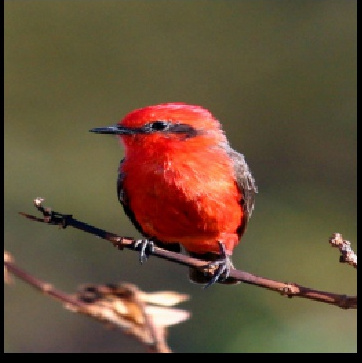}&
\includegraphics[width=1in]{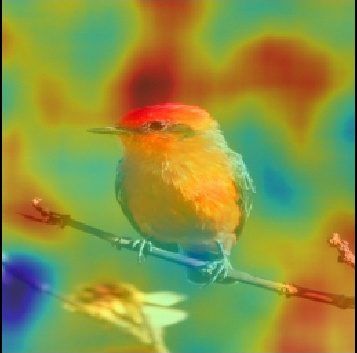}&
\includegraphics[width=1in]{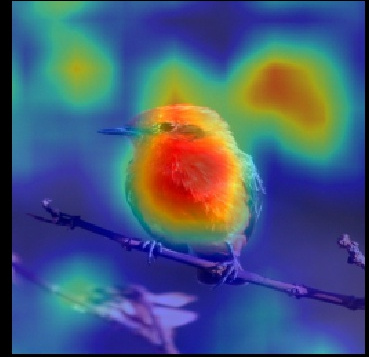} &
\includegraphics[width=1in]{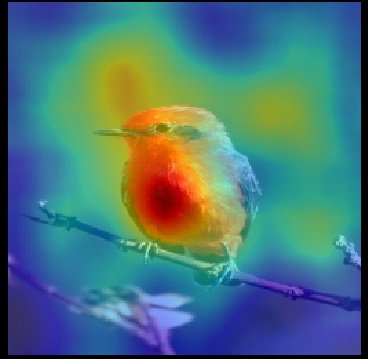} \\

\hline

\end{tabular}

\vspace{2mm}
\caption{\textbf{RISE} heatmaps of images from \textbf{CUB} dataset. In each of the rows the leftmost  column (col 1)  shows  the  original image of the Flycatcher while  col 2-4 show  RISE heatmaps for Vanilla, ALT and AUG Modes respectively (importance increases from blue to red). The column sub-headings denote the corresponding classifier. Note that the heatmaps are generated for the predicted output by the classifiers. Improvement in the explanations can be observed for the ALT and AUG Mode classifiers as compared to the corresponding Vanilla classifier.}
\label{tab:rise_cub}
\end{table*}

\begin{table*}[htbp]
\setlength\tabcolsep{2pt}
\centering
\begin{tabular}{|c|c|c|c|}
\multicolumn{4}{c}{\textbf{GradCAM} heatmaps for \textbf{Horse2Zebra} dataset}\\
\hline
Original Image&
Vanilla Mode &
ALT Mode &
AUG Mode\\
\hline

\includegraphics[width=1in, height=1in]{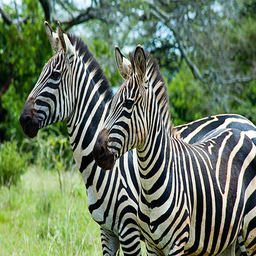} & 
\includegraphics[width=1in]{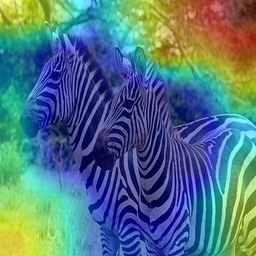}&
\includegraphics[width=1in]{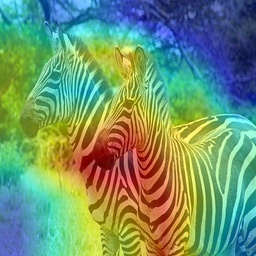}&
\includegraphics[width=1in]{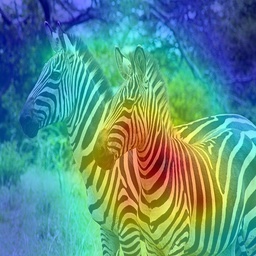}  \\

\includegraphics[width=1in, height=1in]{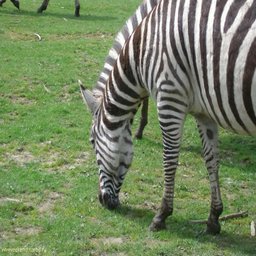} & 
\includegraphics[width=1in]{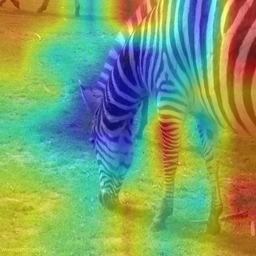}&
\includegraphics[width=1in]{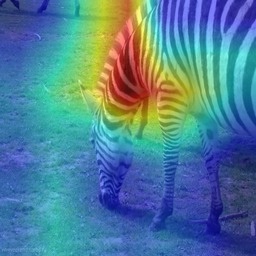}&
\includegraphics[width=1in]{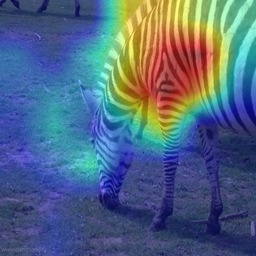}\\

\includegraphics[width=1in, height=1in]{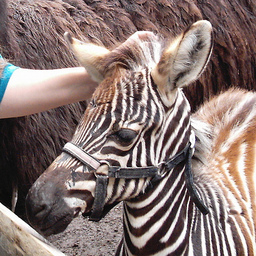} & 
\includegraphics[width=1in]{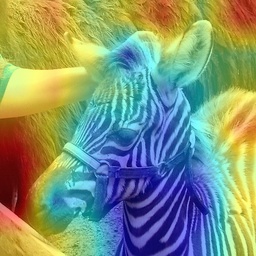}&
\includegraphics[width=1in]{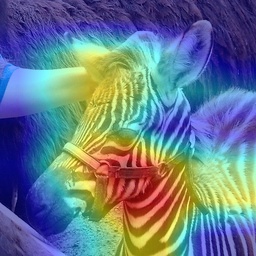}&
\includegraphics[width=1in]{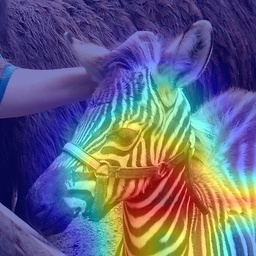}  \\
 
\includegraphics[width=1in, height=1in]{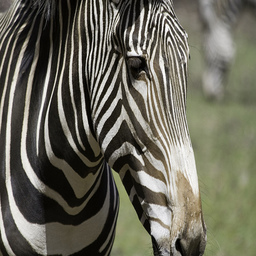} & 
\includegraphics[width=1in]{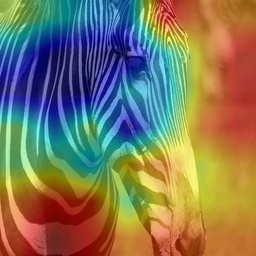}&
\includegraphics[width=1in]{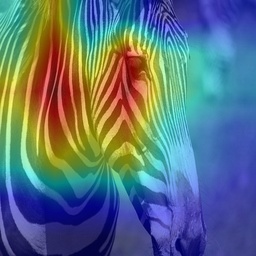}&
\includegraphics[width=1in]{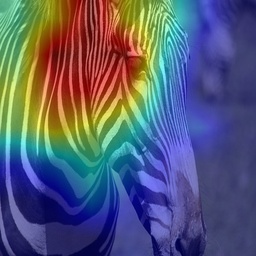} \\

\includegraphics[width=1in, height=1in]{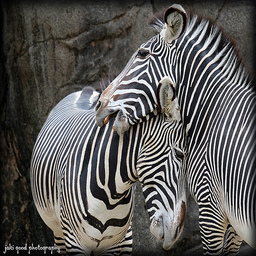} & 
\includegraphics[width=1in]{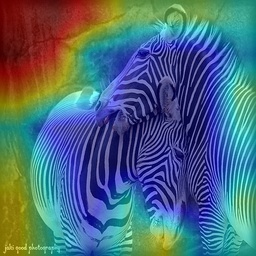}&
\includegraphics[width=1in]{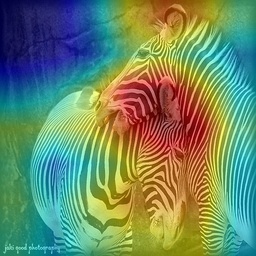}&
\includegraphics[width=1in]{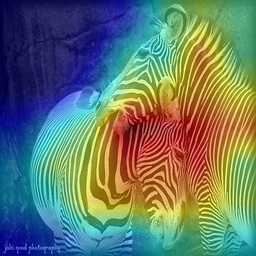}  \\

\hline
\end{tabular}
\vspace{2mm}
\caption{\textbf{GradCAM} heatmaps of images from \textbf{Horse2Zebra} dataset. In each of the rows the leftmost  column (col 1)  shows  the  original image  while  col 2-4 show  GradCAM heatmaps for Vanilla, ALT and AUG Modes respectively (importance increases from blue to red). The column sub-headings denote the corresponding classifier. Note that the heatmaps are generated for the predicted output by the classifiers. Improvement in the explanations can be observed for the ALT and AUG Mode classifiers as compared to the corresponding Vanilla classifier.}
\label{tab:gradcam_H2Z}
\end{table*}

\begin{table*}[htbp]
\setlength\tabcolsep{2pt}
\centering
\begin{tabular}{|c|c|c|c|}
\multicolumn{4}{c}{\textbf{GradCAM} heatmaps for \textbf{CUB} dataset}\\
\hline
Original Image&
Vanilla Mode &
ALT Mode &
AUG Mode\\
\hline

\includegraphics[width=1in, height=1in]{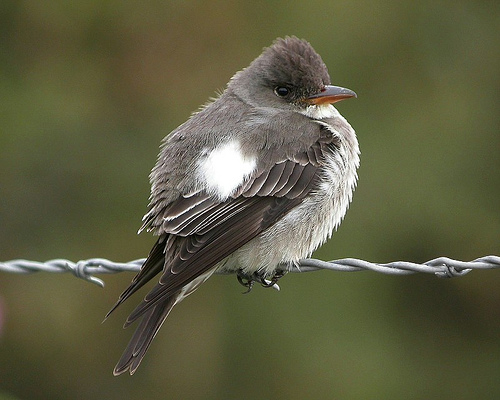} & 
  \includegraphics[width=1in]{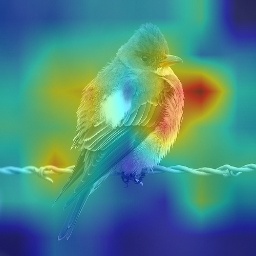}&
 \includegraphics[width=1in]{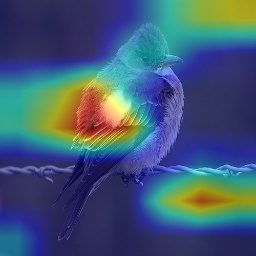} &
 \includegraphics[width=1in]{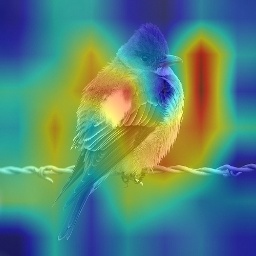}\\
 
 \includegraphics[width=1in, height=1in]{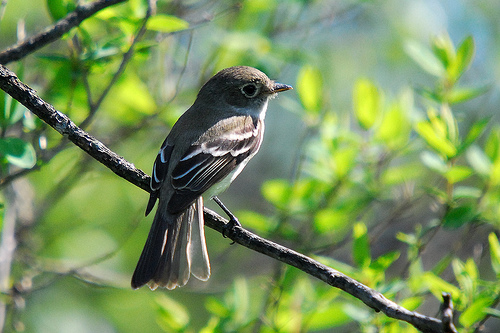} & 
  \includegraphics[width=1in]{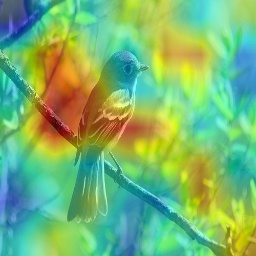}& \includegraphics[width=1in]{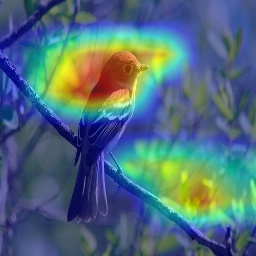}&
   \includegraphics[width=1in]{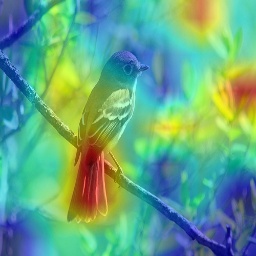}\\
 
 \includegraphics[width=1in, height=1in]{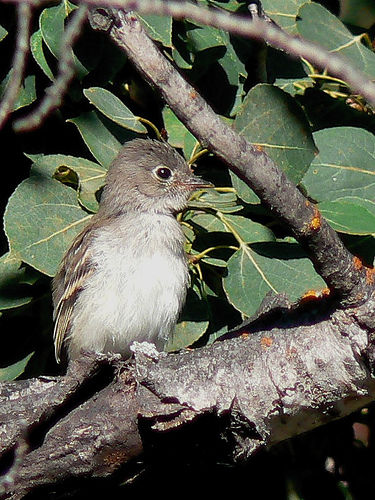} & 
  \includegraphics[width=1in]{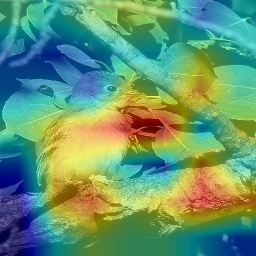}&
  \includegraphics[width=1in]{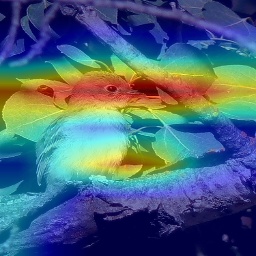} &
 \includegraphics[width=1in]{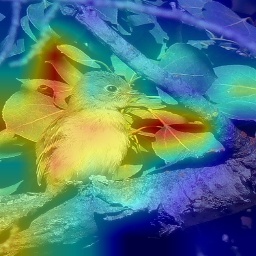} \\
 
 \includegraphics[width=1in, height=1in]{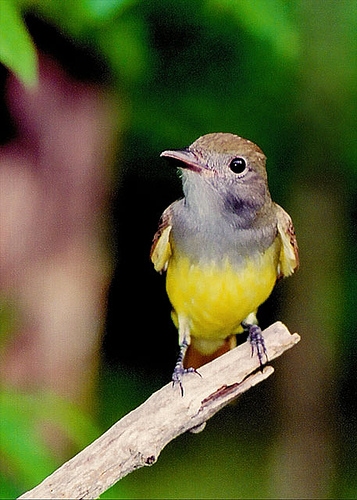} & 
\includegraphics[width=1in]{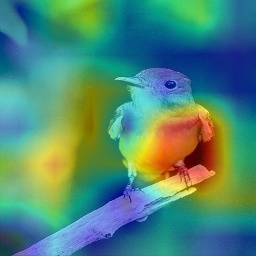}&
\includegraphics[width=1in]{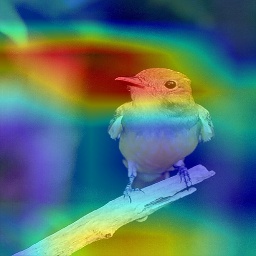} &
\includegraphics[width=1in]{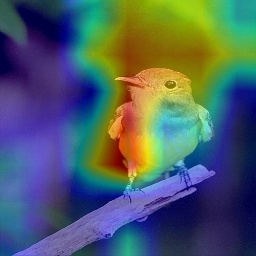}\\

\includegraphics[width=1in, height=1in]{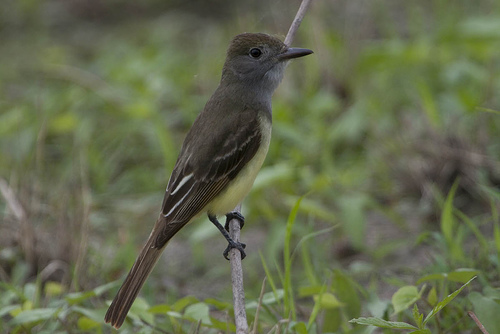} & 
\includegraphics[width=1in]{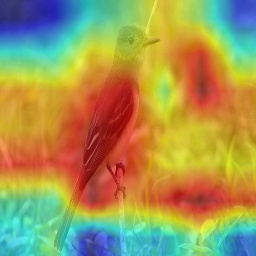}&
\includegraphics[width=1in]{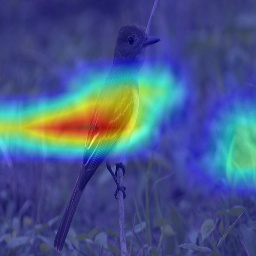} &
\includegraphics[width=1in]{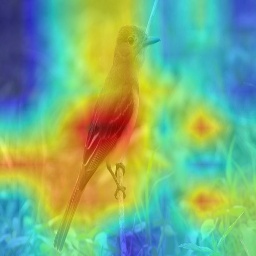} \\

\hline

\end{tabular}
\caption{\textbf{GradCAM} heatmaps of images from \textbf{CUB} dataset. In each of the rows the leftmost  column (col 1)  shows  the  original image  while  col 2-4 show  GradCAM heatmaps for Vanilla, ALT and AUG Modes respectively (importance increases from blue to red). The column sub-headings denote the corresponding classifier. Note that the heatmaps are generated for the predicted output by the classifiers. Improvement in the explanations can be observed for the ALT and AUG Mode classifiers as compared to the corresponding Vanilla classifier.}
\label{tab:gradcam_CUB}
\end{table*}

\end{document}